\documentclass[pmlr,twocolumn,10pt]{jmlr} 

\usepackage{booktabs}
\usepackage{siunitx}

\usepackage[switch]{lineno}

\theorembodyfont{\upshape}
\theoremheaderfont{\scshape}
\theorempostheader{:}
\theoremsep{\newline}

\jmlrvolume{287}
\jmlryear{2025}
\jmlrsubmitted{} 
\jmlrpublished{} 
\jmlrworkshop{Conference on Health, Inference, and Learning (CHIL) 2025}

\usepackage{enumerate}
\usepackage{multirow}
\usepackage{makecell}
\usepackage{wrapfig}
\usepackage{algorithm}
\usepackage{algpseudocode}
\usepackage{caption}
\usepackage{mathabx}

\newcommand{\model}[1]{\texttt{#1}}
\newcommand{\DLinear}{\model{DLinear}}
\newcommand{\MLP}{\model{MLP}}
\newcommand{\TCN}{\model{TCN}}
\newcommand{\TFT}{\model{TFT}}

\newcommand{\Informer}{\model{Informer}}
\newcommand{\PatchTST}{\model{PatchTST}}
\newcommand{\RNN}{\model{RNN}}
\newcommand{\LSTM}{\model{LSTM}}
\newcommand{\NBEATSx}{\model{NBEATSx}}
\newcommand{\NHITS}{\model{NHITS}}

\newcommand{\ETS}{\model{AutoETS}}

\newcommand{\TFTSPercMaxBGTG}{5.4\%}
\newcommand{\TFTSPercMaxBGTGc}{7.7\%}
\newcommand{\TFTSpvalBGTG}{4.6e-4}

\newcommand{\TFTOPercMaxBGTGc}{5.2\%}

\newcommand{\TFTSPercAvgTGSG}{2.4\%}

\newcommand{\TFTSpvalTGSG}{3.6e-12}

\newcommand{\NBEATSXSPercMaxBGTG}{3.7\%} 
\newcommand{\NBEATSXSPercMaxBGTGc}{6.6\%} 
\newcommand{\NBEATSXSpvalBGTG}{7.9e-5} 
\newcommand{\NBEATSXSpvalBGTGc}{5.7e-4}

\newcommand{\NBEATSXOPercMaxBGTGc}{8.5\%} 
\newcommand{\NBEATSXOpvalBGTG}{3.4e-3} 
\newcommand{\NBEATSXOpvalBGTGc}{4.4e-4} 

\newcommand{\NHITSSPercMaxBGTG}{9.2\%} 
\newcommand{\NHITSSPercMaxBGTGc}{16.4\%} 
\newcommand{\NHITSSpvalBGTG}{2.0e-7} 
\newcommand{\NHITSSpvalBGTGc}{3.0e-5}

\newcommand{\NHITSOPercMaxBGTGc}{4.9\%} 
\newcommand{\NHITSOpvalBGTG}{1.3e-3} 
\newcommand{\NHITSOpvalBGTGc}{7.0e-3} 

\newcommand{\NHITSOPercAvgTGTL}{15.8\%}

\newcommand{\NHITSOpvalTGTL}{1.7e-5}

\newcommand{\NHITSOPercAvgTGBL}{14.8\%} 
 
\newcommand{\NHITSOpvalTGBL}{1.7e-6}

\newcommand{\NHITSOPercAvgTGSG}{2.3\%}

\newcommand{\NHITSOpvalTGSG}{4.1e-5}

\title[Global Deep Forecasting with Patient-Specific Pharmacokinetics]{Global Deep Forecasting with Patient-Specific Pharmacokinetics}

\author{%
 \Name{Willa Potosnak} \Email{wpotosna@andrew.cmu.edu}\\
 \addr Auton Lab, School of Computer Science, Carnegie Mellon University \\
 \Name{Cristian Challu} \Email{cristian@nixtla.io}\\
 \addr Nixtla \\
 \Name{Kin G. Olivares} \Email{kigutie@amazon.com}\\
 \addr Amazon \\
 \Name{Keith A. Dufendach} \Email{dufendachka@upmc.edu}\\
 \addr University of Pittsburgh Medical Center, Division of Cardiac Surgery \\
 \Name{Artur Dubrawski} \Email{awd@andrew.cmu.edu}\\
 \addr Auton Lab, School of Computer Science, Carnegie Mellon University \\
}

\begin{document}

\maketitle

\begin{abstract}
Forecasting healthcare time series data is vital for early detection of adverse outcomes and patient monitoring. However, it can be challenging in practice due to variable medication administration and unique pharmacokinetic (PK) properties of each patient. To address these challenges, we propose a novel hybrid global-local architecture and a PK encoder that informs deep learning models of patient-specific treatment effects. We showcase the efficacy of our approach in achieving significant accuracy gains in a blood glucose forecasting task using both realistically simulated and real-world data. Our PK encoder surpasses baselines by up to \NHITSSPercMaxBGTGc\ on simulated data and \NHITSOPercMaxBGTGc\ on real-world data for individual patients during critical events of severely high and low glucose levels. Furthermore, our proposed hybrid global-local architecture outperforms patient-specific PK models by \NHITSOPercAvgTGTL, on average.
\end{abstract}

\paragraph*{Data and Code Availability}
We use open-source simulated and real-world data that is available to other researchers. The simulated data includes an open-source Python implementation of the FDA-approved UVa/Padova Simulator (2008 version), which is contained in the `simglucose' GitHub repository \citep{simglucose_dataset, visentin2016towards_single_day_simulator}. The real-world data includes the OhioT1DM 2018 and 2020 datasets \citep{ohiot1dm_dataset}. Information on accessing this dataset, including a link to the required Data Use Agreement (DUA), is provided in the paper by \cite{ohiot1dm_dataset}. Implementations of models used in our work were obtained from the open-source \texttt{Neuralforecast}~\citep{olivares2022library_neuralforecast} and \texttt{StatsForecast}~\citep{garza2022statsforecast} libraries. We implemented our method by adopting the \texttt{Neuralforecast} repository to leverage their model training framework. Our research code, along with the \texttt{Neuralforecast} models used in our research, is available at \url{https://github.com/PotosnakW/neuralforecast/tree/pk_paper_code}.

\paragraph{Institutional Review Board (IRB)} Our research does not require IRB approval.

\section{Introduction}
\label{section:introduction}
\begin{figure}[t!]
    \centering
    \includegraphics[width=0.5\textwidth]{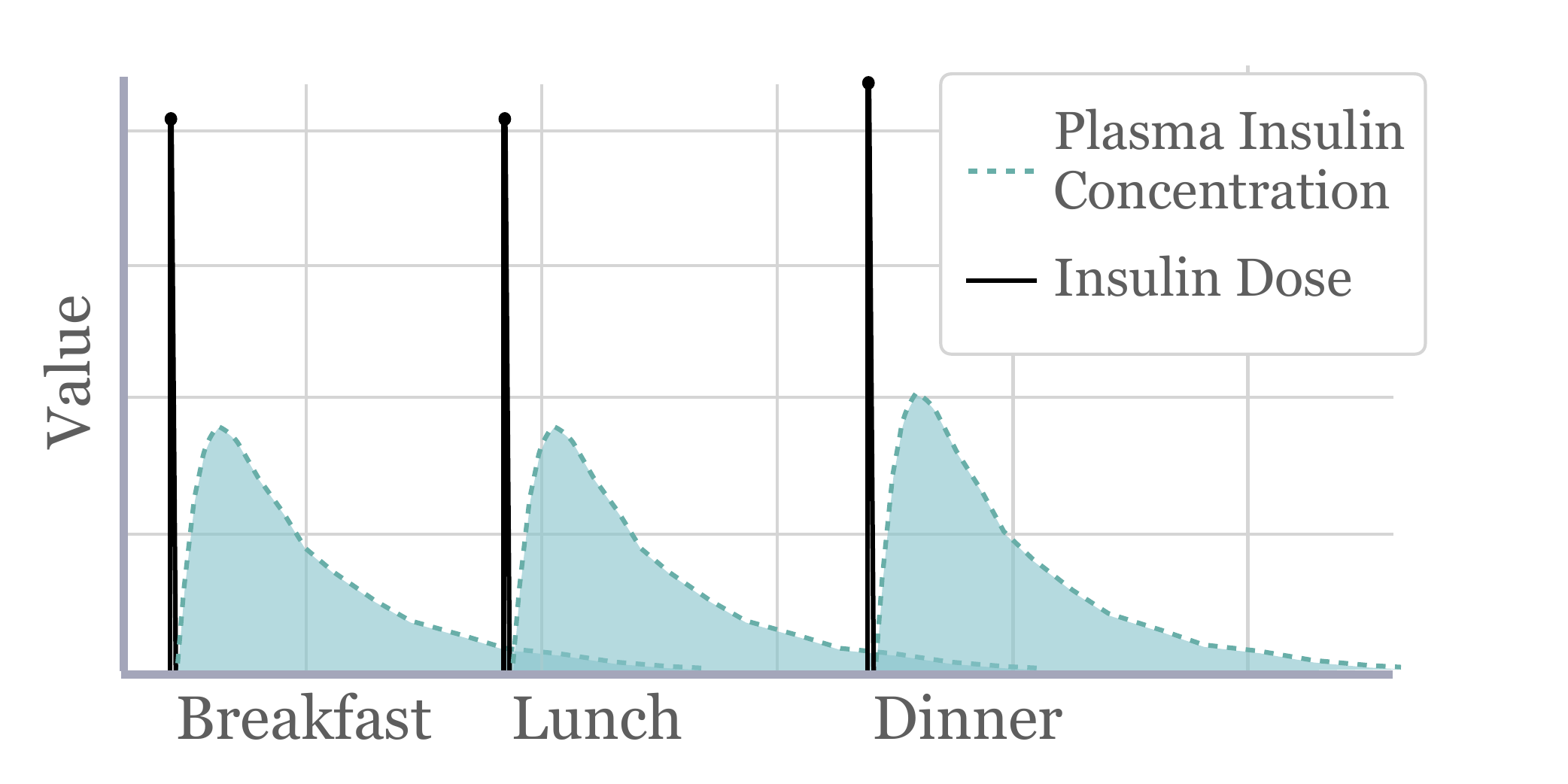}
    \caption{Insulin dose is recorded, but plasma insulin concentration is not directly measurable and typically requires invasive procedures and laboratory tests.} 
    \label{fig:motivation}
\end{figure}

Leveraging the predictive power of exogenous variables presents unique challenges, especially in healthcare, where the frequency of patient signals and exogenous factors, such as medication doses or treatments, often exhibit a mismatch in temporal resolution. Furthermore, while the drug dose can be observed and recorded, the body's interaction with the drug cannot be directly observed and is typically inferred from its plasma concentration, as illustrated in Fig.~\ref{fig:motivation}. Consequently, the recorded drug dose alone cannot capture time-dependent pharmacokinetics, including processes such as absorption and elimination.

Consider the task of blood glucose monitoring for patients with Type-1 Diabetes Mellitus (T1DM) or insulin-dependent T2DM patients who must regularly monitor and regulate their blood glucose levels. Careful and prompt administration of insulin and intake of carbohydrates is required to prevent hyper- and hypoglycemic events, which account for numerous hospital admissions annually, and can cause irreversible organ damage or, in severe cases, be fatal. A 2016 report estimated that hypoglycemia causes approximately 100,000 emergency department (ED) visits per year due to insulin-related errors, with nearly one-third resulting in hospitalization \citep{hypoglycemia2014geller}. Another study reports that in 2014, there were 184,255 ED visits and in-patient U.S. hospital admissions for diabetic ketoacidosis (DKA), often associated with hyperglycemia, with 70.6\% occurring in T1DM patients \citep{dka2020benoit}. Another study reported a trend toward increasing hospitalizations for DKA in T1DM patients from 2008 to 2018 \citep{dka2021kchloo}.

For healthcare challenges like this, advanced forecasting technologies can help patients and clinicians monitor, anticipate, and proactively address emerging abnormalities to prevent adverse events. However, accurate forecasting requires models that can leverage exogenous information, such as treatments, while capturing latent treatment kinetics and their effects to generate reliable forecast trajectories. To address this, we propose a hybrid model that integrates cohort-level information with patient-specific pharmacokinetics to improve the prediction of blood glucose responses to treatment.

The main contributions of this paper are:
\begin{enumerate}[(i)]
    \item \textbf{Hybrid Global-Local Architecture.} We bolster global architectures (with parameters shared across patients) with a learnable, patient-specific encoder.
    \item \textbf{Pharmacokinetic (PK) Encoder.} We introduce a novel deep learning model-agnostic module that generates plasma concentration profiles to capture time-dependent treatment effects.
    \item \textbf{Multiple Dose Effects Framework.} We provide a computationally efficient framework for encoding the effects of multiple doses. We verify this approach mathematically under the assumption of linear pharmacokinetics.
    \item \textbf{State-of-the-Art Results.} We obtain significant accuracy improvements on large-scale simulated and real-world patient blood glucose forecasting datasets.
\end{enumerate}

\subsection{Pharmacokinetics Background}\label{section:background}

\begin{figure}[t!]
    \centering
    \includegraphics[width=0.45\textwidth]{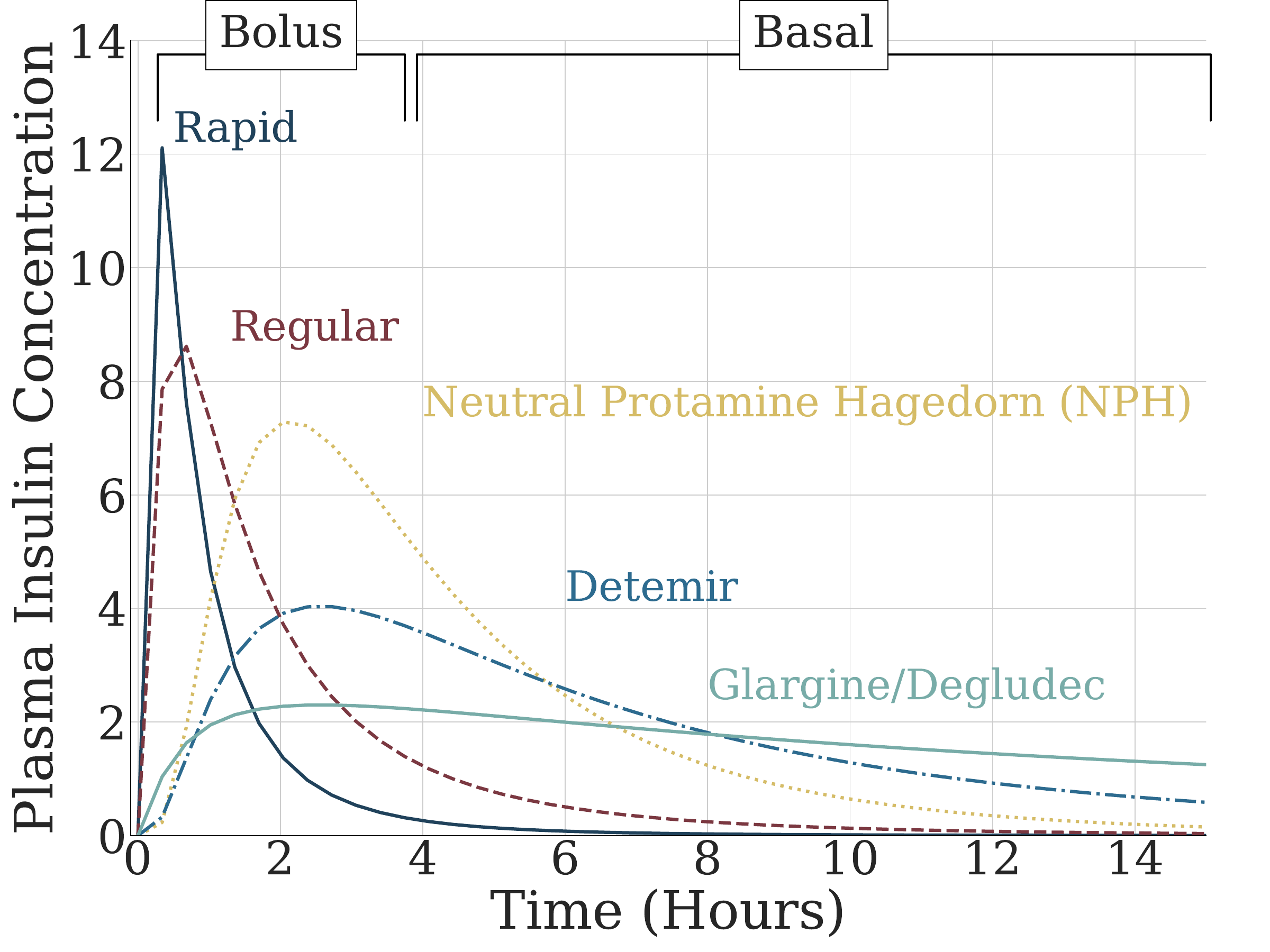}
    \caption{Pharmacokinetic models of plasma insulin concentration over time for various insulin types.}
    \label{fig:bolus_basal_diagram}
\end{figure}

Pharmacokinetic (PK) modeling is a natural solution to the aforementioned problem, as it can reveal medication effects over time through well-established knowledge of changes in plasma insulin concentration driven by absorption and elimination characteristics ~\citep{binder1984insulin}. Empirical and invasive studies show that concentration-time profiles for subcutaneous injections follow right-skewed curves. The extent of skew varies with insulin type and patient-specific factors, as shown in Fig.~\ref{fig:bolus_basal_diagram}, where the area under the concentration curve measures drug \textit{bioavailability}, or the fraction of the absorbed dose that reaches the intended site of systemic circulation intact. Most drugs used in clinical practice are assumed to follow linear PK \citep{basic_pharmacokinetics}. Under this model, the area under the concentration curve is directly proportional to the drug dose. A complete reference for supporting PK information regarding concentration-time profiles, insulin analogs, bioavailability, first-order kinetics, and linear PK can be found in Appendix~\ref{apd:pk_background}.

\vspace{-10pt}
\section{Related Work}

\textbf{Blood Glucose Forecasting.} Neural Forecasting has previously been used to predict blood glucose levels, outperforming traditional statistical methods. RNNs and LSTMs have been used both as models and as baselines in numerous studies \citep{qian2021exportodes, li2019convolutional, fox2018deep, rabby2021stacked, rubin2020deep, rubin_falcone2022glucose_sparse, glucose2020wild}. Similarly, CNNs \citep{xie2020glucose_benchmarking, li2019convolutional, li2019glucose_glunet} and MLPs \citep{mcshinsky2020comparison, glucose2020wild, jahangir2017expert} have been used in proposed models and as baselines.

\paragraph{Modeling Blood Glucose Treatment Effects.} Prior work has looked at modeling treatment effects on blood glucose levels. Miller et al.\ introduced one of the first hybrid statistical-physiological models, the Deep T1D Simulator (DTD-Sim), which integrates glucose-insulin dynamics with deep learning through a deep state-space model \citep{glucose2020wild}. \citet{odnoblyudova2023nonparametric} improved the prediction of blood glucose using multi-output Gaussian processes (MOGP) to model the composite effects of meal components, such as carbohydrates and fats. Unlike meals, insulin is a homogeneous protein hormone, making component-based modeling inapplicable to our task. Instead, our objective is to capture the holistic patient-specific effects of the drug and the composite effects of multiple doses. 

Despite advances in treatment effect modeling, current approaches face limitations, including (1) reliance on individualized, subject-specific models and (2) inflexible architectures that hinder adaptability to various base models, forecasting tasks, and model inputs. The next paragraph outlines these limitations and how our proposed contributions address them.

Previous work has focused on developing individualized networks, often overlooking the advantages of global models that learn from multiple, diverse time series—a technique known as cross-learning \citep{smyl2019esrnn, spiliotis2021cross_learning}, and employed by Time Series Foundation Models (TFSMs) \citep{goswami2024moment}. Our hybrid global-local architecture enables cohort-level training to develop a global model applicable to multiple patients while also preserving patient-specific information for personalized forecasts. Regarding architectures, models such as the UVA / Padova simulator depend on predefined ODEs for glucose dynamics, restricting adaptability to other targets and exogenous inputs \citep{visentin2016towards_single_day_simulator, glucose2020wild}. Similarly, deep learning approaches for sparse exogenous variables often rely on fixed base models like LSTMs \citep{rubin_falcone2022glucose_sparse}. Our proposed contributions support diverse deep learning models, and are flexible to forecasting targets and input variables, ensuring future adaptability as more advanced tools emerge. Other prior work that proposes the integration of PK knowledge with deep learning focuses on domains such as long-term disease progression in cancer and COVID-19, where such PK models are not directly applicable to blood glucose and insulin kinetics. \citep{hussain2021pharmacokinetics_state_space, qian2021exportodes}. Our study provides a formal framework for integrating PK in deep learning under linear PK assumptions that are applicable to many drugs used in clinical practice \citep{basic_pharmacokinetics}.

\label{section:literature}
\section{Methods}

We aim to forecast blood glucose values 30 minutes into the future consistent with prior work \citep{rubin_falcone2022glucose_sparse, xie2020glucose_benchmarking, li2019glucose_glunet}, based on a 10-hour history of blood glucose, carbohydrate (CHO), insulin basal, and insulin bolus values. We choose a 10-hour duration to balance sufficient historical information with our GPU memory constraints. Moreover, the duration of rapid-acting insulin (Humalog and Novolog) used in insulin pumps by patients in our cohort typically ranges from 2 to 5 hours \citep{insulin_duration_cc, insulin_duration_ada}. Thus, a 10-hour time span fully captures multiple dose durations, enabling the model to learn treatment effects across various dosages and durations. We also demonstrate that the significant performance improvement of the proposed methods holds across other various lag times in Appendix~\ref{apd:contextlen_ablation}.

\subsection{PK Encoder}\label{section:pk_encoder}

\begin{figure}[t!]
    \centering
    \includegraphics[width=0.495\textwidth]{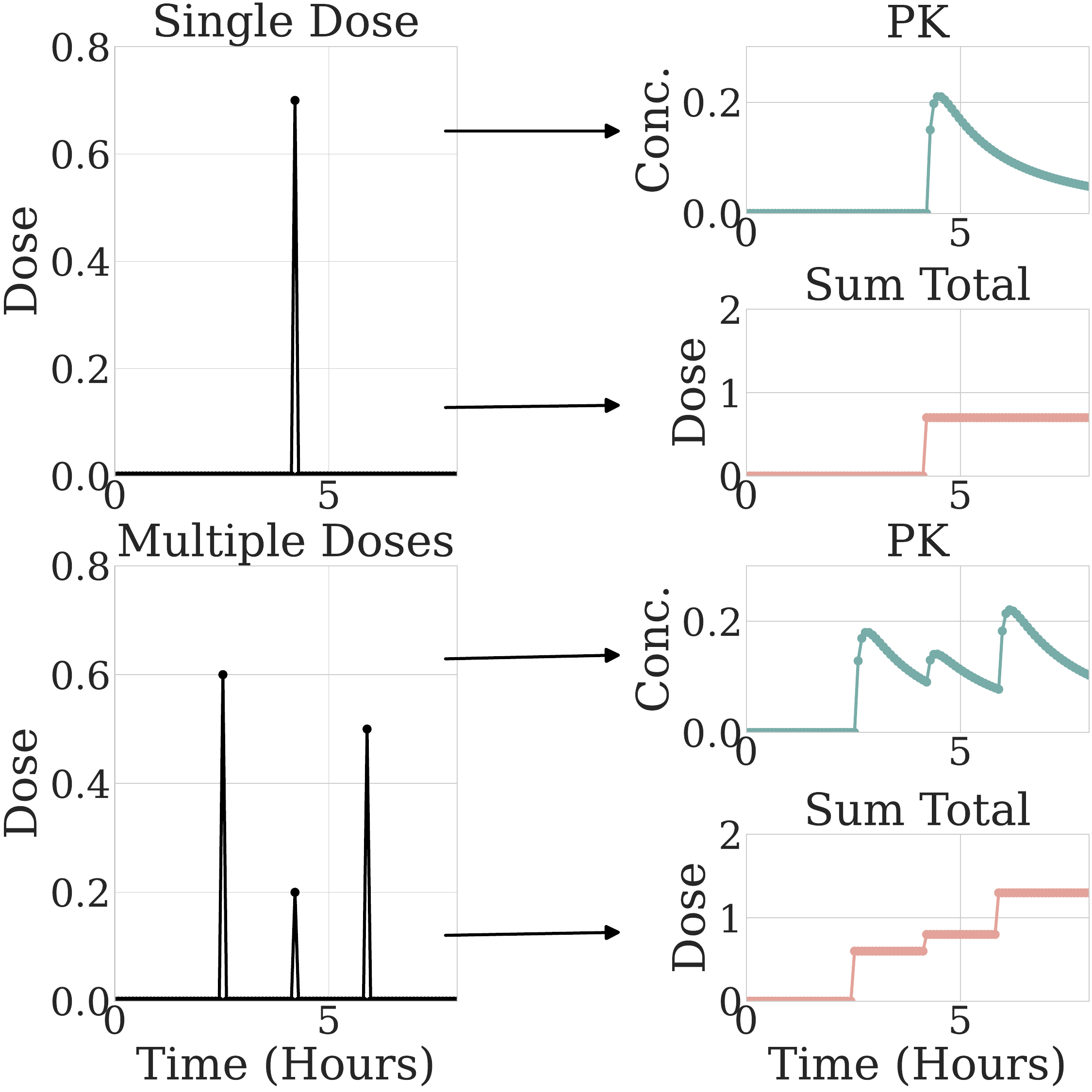}
    \caption{Medication administrations are represented as sparse variables in time series data. We propose a pharmacokinetic (PK) encoder to effectively capture time-dependent plasma drug concentration (Conc.).}
    \label{fig:insulin_featurization}
\end{figure}

Medication administrations are recorded as instantaneous dosing events, leading to sparse observations in time series data. We propose a novel PK encoder that informs deep learning models of time-dependent plasma drug concentrations to enable more accurate forecasting of treatment outcomes as shown in Fig.~\ref{fig:insulin_featurization}. The encoder consists of a function $C$ that generates a concentration-time profile for a single specified dose $x$ at time $t$ using a PK parameter $k$ affecting elimination rate:
\begin{align}
    C(t, x, k) = \frac{x}{tk\sqrt{2\pi}} \exp\{-\frac{1}{2k^2}(\log(t) - 1)^2\}.
\end{align}
Here, $C$ is equivalent to the log-normal probability density function with $\sigma=k$ and $\mu = 1$ and a scaling factor $x$. The intuition for the log-normal shape of the concentration curves stems from PK literature, as outlined in Section~\ref{section:background} and Appendix~\ref{apd:pk_background}.

Under linear PK assumptions, the PK parameters, including bioavailability, are assumed to be constant with respect to the drug. Given this assumption, we show mathematically that the area under the concentration curve is bounded by the dose in Proposition~\ref{proposition:auc_bound} in Appendix~\ref{apd:first_proposition}. To model bioavailability and incorporate dose information directly into the concentration curves, we use the dose $x$ as a scaling factor in $C$. This ensures that, under the simplifying assumption of 100\% bioavailability, the area under the curve integrates to the dose. Scaling the concentration curve in this way affects its magnitude but does not alter PK parameters that determine its shape.

Let $N$ be the number of subjects in the dataset. The PK parameter vector $\mathbf{k} \in \mathbb{R}^{N \times 1}$ represents patient-specific embedding weights that control the shape of the patient-specific concentration curves, where $\mathbf{k}^{(i)}$ corresponds to the PK parameter $k$ for the $i$-th patient. The vector $\mathbf{k}$ is initialized with values between 1 and 2 prior to model training, with the initial value treated as a tunable hyperparameter.

\begin{figure*}[ht!]
\centering
\includegraphics[width=1.0\textwidth, trim=10 75 10 2, clip]{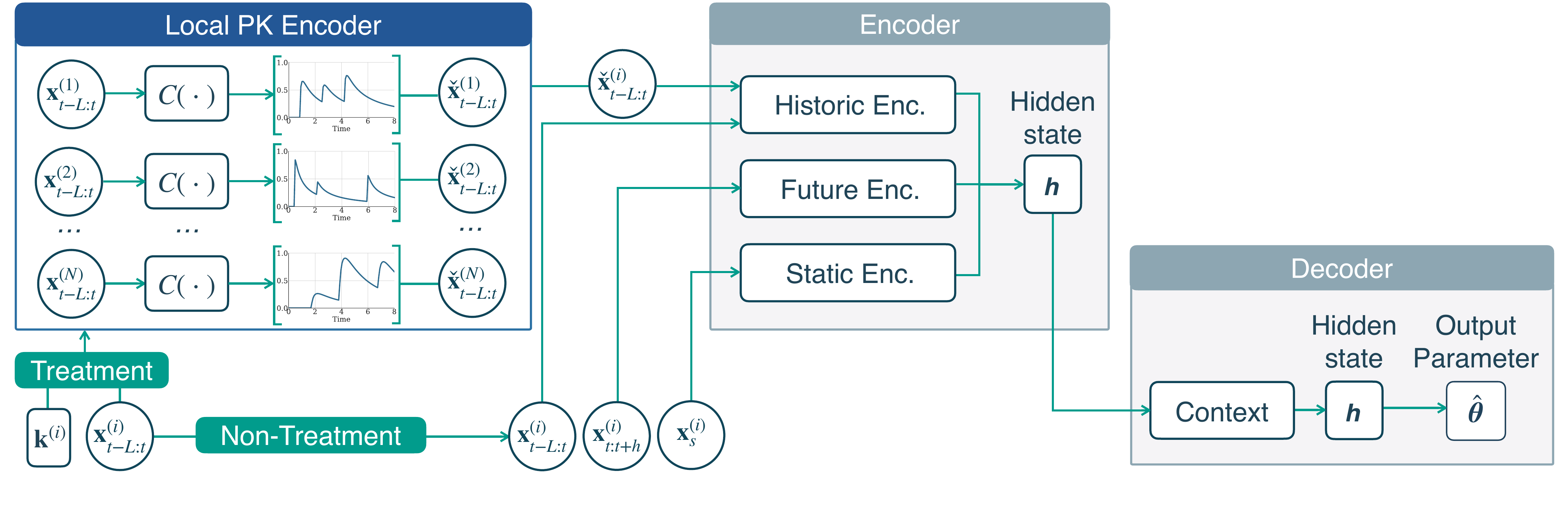}
\caption{Our hybrid global-local architecture combines global model parameters shared across patients with patient-specific pharmacokinetic (PK) parameters. Sparse treatment time series features, such as medication doses, are input into the PK encoder to generate patient-specific treatment concentration curves $\check{\textbf{x}}$. PK encoder output $\check{\textbf{x}}$ and any other exogenous features $\textbf{x}$ are used as deep learning model inputs. During each training step, the model learns relevant hidden states and output parameters $\hat{\theta}$ across patients, while $\textbf{k}^{(i)}$ values are updated for each patient $i$ using gradient descent optimization.}
\label{fig:method}
\end{figure*}

The encoder is trained end to end with the model. During training, a sparse treatment feature $\mathbf{x}^{(i)} \in \mathbb{R}^{1 \times L}$, representing dose administrations within an input window of length $L$ up to and including the current time $T$, is passed to the PK encoder along with the corresponding parameter $\mathbf{k}^{(i)}$ to generate a concentration curve feature $\widecheck{\mathbf{x}}^{(i)} \in \mathbb{R}^{1 \times L}$, as shown in Fig.~\ref{fig:method}. In the \textit{single-dose setting}, where a dose is recorded at time step $t_d \in \{T-L+1, \dots, T\}$, the encoder replaces the sparse treatment feature $\mathbf{x}_{T-L+1:T}^{(i)}$ with the corresponding concentration feature $\widecheck{\mathbf{x}}_{T-L+1:T}^{(i)}$, computed as:
\begin{align}
\widecheck{\mathbf{x}}^{(i)}_t &= 
\begin{cases}
0, & \!\text{for } t \in [T-L+1, t_d) \\
C\left( \nu(t - t_d), \mathbf{x}^{(i)}_{t_d}, \mathbf{k}^{(i)} \right), &\!\text{for } t \in [t_d, T]
\end{cases}
\end{align}
where $t$ indexes time steps within the window, $\nu$ is the sampling interval of the data, and $C(\cdot)$ computes the concentration at elapsed time $\nu(t - t_d)$ since the dose was administered. During training, the parameter $\mathbf{k}^{(i)}$ is updated for each patient $i$ via gradient descent.

Forecasts $\hat{\textbf{y}}$ are generated by the model $f_\theta$ for each time step within the forecast horizon $H$ as a function of historical glucose values $\textbf{y}$, treatment concentration curves $\widecheck{\textbf{x}}$, and static features $\textbf{s}$, recorded within the input window up to and including time $T$, 
\begin{align}
\hat{\textbf{y}}^{(i)}_{T+1:T+H} &= f_\theta(\textbf{y}^{(i)}_{T-L+1:T}, \widecheck{\textbf{x}}^{(i)}_{T-L+1:T}, \textbf{s}^{(i)}).
\end{align}
In the proposed context,  $\textbf{s}$ includes patient age, weight, one-hot encoded subject identification number, and insulin pump type. $\widecheck{\textbf{x}}_{T-L+1:T}$ includes concentration curve values for basal insulin $\widecheck{\textbf{x}}_{\text{basal}}$, bolus insulin $\widecheck{\textbf{x}}_{\text{bolus}}$, and CHO $\widecheck{\textbf{x}}_{\text{CHO}}$. We include treatment effects of carbohydrates because, similar to drugs, they undergo absorption and metabolism processes, resulting in glucose and other sugars entering the bloodstream. 

\subsection{Modeling Multiple Dose Effects}
Multiple insulin doses may occur at close intervals in an event often referred to as ``insulin stacking'' as shown in Fig.~\ref{fig:insulin_featurization} and Fig.~\ref{fig:insulin_stacking}. To model PK effects for multiple doses within an input window, we generate a concentration curve for each dose event and sum the concentration curves within the window. Under the assumption of linear PK (outlined in~\ref{section:background} and Appendix~\ref{apd:pk_background}), we show that the overall effect, or concentration, of $n$ doses of the same drug is equal to the sum of the individual effects which is verified mathematically in Proposition~\ref{proposition:concentration_summation} in Appendix~\ref{apd:first_proposition}. To efficiently generate concentration curves for multiple doses within an input window, the PK encoder uses a matrix $\textbf{w} \in \mathbb{R}^{L \times L}$ to support vectorized computation at each timestamp containing a dose, within the lag time $L$. This approach enables efficient modeling of multiple effects in $\mathcal{O}(1)$ time. Each row $t$ of $\textbf{w}$ represents the elapsed time steps until the end of the window, with the time count shifting one step to the right in each subsequent row. A concentration curve is generated for each row of $\textbf{w}$ using the dose at the corresponding time index. If no dose is administered at a particular time $t$, the corresponding row of $\textbf{w}$ is zeroed out. The final concentration curve feature, which captures the combined impact of multiple doses, is obtained by aggregating the individual curves across the rows:
\begin{align} 
    \mathbf{w} &=
    \begin{bmatrix}
    0 & 1 & 2 & 3 & \cdots & L{-}1 \\
    0 & 0 & 1 & 2 & \cdots & L{-}2 \\
    \vdots & \vdots & \vdots & \vdots & \ddots & \vdots \\
    0 & 0 & 0 & 0 & \cdots & 0
    \end{bmatrix}
    \in \mathbb{R}^{L \times L} \\
    \widecheck{\textbf{x}}^{(i)} &= \sum_{t=0}^{L-1} C(\nu\textbf{w}_{t, :}, \textbf{x}^{(i)}_t, \textbf{k}^{(i)}) 
\end{align}

\noindent Here, $\nu$ is the sampling interval of the data, $\textbf{x}^{(i)} \in \mathbb{R}^{1 \times L}$ is a sparse exogenous time series treatment feature, and $\widecheck{\textbf{x}}^{(i)} \in \mathbb{R}^{1 \times L}$ is the new concentration curve feature for patient $i$. 

The PK encoder training algorithm~\ref{alg:pk_model_training} is described in Appendix~\ref{apd:first_algorithm}, which can be integrated into any deep learning architecture capable of handling exogenous variables in addition to modeling the target variable.

\subsection{Modeling Dose-Dependent Effects}\label{section:methods_dose_dependent_effects}

A majority of drugs are assumed to adhere to linear pharmacokinetics (PK), where a linear relationship is expected between drug dose and its concentration in the body \citep{basic_pharmacokinetics}. In practice, however, whether a drug’s PK parameters follow linear or nonlinear behavior can vary based on factors such as the drug’s physicochemical properties, the route of administration, and the injection site. A key characteristic of the linear PK assumption is that PK parameters remain constant and do not change with variables such as dose or time. Dose- or time-dependent nonlinearity occurs when PK parameters change nonproportionally with dose or time, respectively. In dose-dependent nonlinear PK, the relationship between drug dose and body concentration is no longer linear, resulting in dose-related changes in PK parameters. While insulin is not expected to exhibit classic time-dependent nonlinearity, it may demonstrate dose-dependent nonlinearity. For instance, subcutaneous injections are administered into the subcutaneous tissue, where higher doses can saturate elimination pathways, causing slower absorption rates \citep{soeborg2009absorption, gradel2018absorptionfactors}. When PK parameters are dose-dependent, the resulting plasma drug concentration profiles also vary with dose.

Insulin pumps administer the same type of insulin (rapid-acting) for the basal and bolus doses. However, bolus insulin doses are generally much larger than basal doses as highlighted in Figs.~\ref{fig:simglucose_insulin_kde} and~\ref{fig:ohiot1dm_insulin_kde} in Appendix~\ref{apd:first_data}. To account for potential dose-dependent effects, we model basal and bolus doses separately by learning distinct PK parameters, $\mathbf{k}_{\text{basal}}$ and $\mathbf{k}_{\text{bolus}}$, allowing us to capture PK variations without resorting to a fully nonlinear model. This approach maintains linear assumptions within each dose type, simplifying the modeling process while still capturing differences between low-dose (basal) and high-dose (bolus) PK. For a single dose of basal insulin $\mathbf{x}_{\text{basal}}$ or bolus insulin $\mathbf{x}_{\text{bolus}}$, administered at time step $t_d$, we compute the plasma insulin concentration at time $t \geq t_d$ for patient $i$:
\begin{align}
    \widecheck{\textbf{x}}^{(i)}_{\text{basal}} &= C\left(\nu(t-t_d), \textbf{x}^{(i)}_{t_d,\text{basal}}, \textbf{k}_{\text{basal}}^{(i)}\right),\\
    \widecheck{\textbf{x}}^{(i)}_{\text{bolus}} &= C\left(\nu(t-t_d), \textbf{x}^{(i)}_{t_d,\text{bolus}}, \textbf{k}_{\text{bolus}}^{(i)}\right).
\end{align}

Treating each dose category as a distinct entity is an approximation and may not capture all nuances of dose-dependent nonlinear PK. All physiological models are approximations of complex biology. Our goal is to evaluate a novel deep learning-based PK modeling approach based on established linear PK knowledge, while accounting for potential dose-dependent effects, balancing simplicity and precision. This framework also ensures broader applicability across various medications, as most drugs are assumed to follow linear PK \citep{basic_pharmacokinetics}.

The equations presented are shown in a patient-specific manner to highlight that the PK encoder generates individualized information. However, these equations can be applied to produce patient-specific outputs for multiple patients simultaneously using vectorized operations.

\subsection{Models}
To assess forecasting performance across various model architectures, we trained global models with 11 different algorithms as baselines. Additionally, to assess our proposed hybrid global-local architecture, we compare the best performing model with the PK encoder to both purely local and global models.

\paragraph{Statistical and Deep Learning Baselines.} We employed Exponential Smoothing (\ETS; \citealt{ets_2008}) as our statistical baseline model. Our deep learning baselines include models from various architectural categories, such as Linear models, Multi-Layer Perceptron (MLP)-based, Recurrent Neural Network (RNN)-based, Convolutional Neural Network (CNN)-based, and Transformer-based models. We employ the following deep learning models: \DLinear~\citep{zeng_2023_dlinear}, \MLP~\citep{rosenblatt1958_mlp}, \NHITS~\citep{challu_olivares2022_nhits}, \NBEATSx~\citep{oreshkin2020nbeats, OlivaresChallu2022_nbeats}, \RNN~\citep{elman1990_rnn, cho2014_rnn_encoder_decoder}, Long-Short-Term Memory (\LSTM; \cite{sak2014_lstm}), Temporal Convolutional Network with an \MLP\ decoder (\TCN; \cite{bai2018_tcn, oord2016_tcn}), \Informer~\citep{zhou2021informerefficienttransformerlong}, patch time series Transformer (\PatchTST; \cite{nie2023patchtst}), and Temporal Fusion Transformer (\TFT; \cite{lim2021_tft}). More information on these models and related prior work is provided in Appendix~\ref{apd:first_models}. Details on model training and hyperparameter selection are provided in Appendix~\ref{apd:first_hyperparameters}.

\paragraph{Global Model Baselines.} Global models are trained on all patient data to create a single model. Global model baselines include: (1) models trained with only the blood glucose signal and static features, (2) models trained with blood glucose signals, static features, and sparse historical exogenous features (CHO and insulin dose values), and (3) models trained with blood glucose signals, static features, and sparse historical exogenous features transformed using the ``Sum Total" (ST) approach, which converts features into a cumulative sum of values across an input window \citep{rubin_falcone2022glucose_sparse}. We include models trained without exogenous variables as baselines to show the impact of incorporating sparse exogenous information for each model. Together with models trained on sparse and Sum Total features, these baselines demonstrate the importance of effective transformations—such as our PK encoder—in reducing forecast error.

With the exception of \TFT, most Transformer-based models, including \Informer\ and \PatchTST, are not designed to model covariates. For these two models, we can only report global model results for those trained on univariate blood glucose data. Since our proposed PK encoder is intended to supplement model architectures capable of incorporating exogenous variables alongside the endogenous signal, we focus on evaluating the PK encoder in top-performing models that are capable of modeling covariates, such as \TFT, \NBEATSx, and \NHITS. Beyond modeling covariates, residualized MLP-based models like \NBEATSx~and \NHITS~are valuable to demonstrate the utility of our PK encoder as \NHITS~has achieved SOTA performance, outperforming RNNs and Transformers in long-range forecasting tasks while avoiding the quadratic complexity of self-attention in Transformer-based architectures \citep{challu_olivares2022_nhits, aws2024chronos}. 

\paragraph{Local Model Baselines.} Patient-specific models using sparse exogenous inputs, as well as those incorporating the PK encoder, were compared against our hybrid global-local PK encoder model. These baselines highlight the effectiveness of hybrid global-local architectures that leverage cross-learning from multiple time series.

\subsection{Model Training} Models were trained using \texttt{Adam} optimizer \citep{2017_adam_optimizer}. Optimal hyperparameters were selected via cross-validation on a validation set by minimizing Huber loss across all prediction time points. Given the use of forward-filling to account for missing values in the OhioT1DM dataset, timestamps with forward-filled values were omitted from the validation set.  Huber estimator was selected as the loss function, $\mathcal{L}$, given its ability to reduce the impact of outliers and offer more stable results compared to other functions:
\begin{equation}
\mathcal{L}(\textbf{y}, \hat{\textbf{y}}, \delta) = 
\left\{
\begin{array}{ll}
\frac{1}{2}(\textbf{y} - \hat{\textbf{y}})^2, &\text{if } |\textbf{y} - \hat{\textbf{y}}| \leq \delta, \\
\delta (|\textbf{y} - \hat{\textbf{y}}| - \frac{1}{2}\delta), &\text{otherwise}.
\end{array}
\right.
\end{equation}
More information on model training and hyperparameters in provided in Appendix~\ref{apd:first_hyperparameters}.

\subsection{Model Evaluation}
Trained models were used to generate rolling window forecasts at one-step intervals. Similar to prior work \citep{rubin_falcone2022glucose_sparse, xie2020glucose_benchmarking}, we evaluate forecasts for each test set in the two datasets using two metrics: mean absolute error (MAE) and root mean squared error (RMSE). We compute results across all points within the forecast horizon, and present the average results over 8 trials of individually trained models. We also evaluate the performance of the models exclusively at time points where blood glucose levels reach or fall below critical thresholds relevant to hypoglycemic and hyperglycemic events (i.e., $\leq$70 mg/dL and $\geq$180 mg/dL). For the OhioT1DM dataset, timestamps with forward-filled values were omitted from evaluation. To evaluate whether hybrid global-local PK models significantly outperform baselines, we use paired t-tests computed on the patients' mean MAE results. Metric formulations and statistical tests are detailed in Appendix~\ref{apd:evaluation}.

\label{section:methodology}
\section{Cohort}

We use simulated and real-world datasets containing patient insulin, CHO time series, and blood glucose measurements. These open-source datasets ensure reproducibility and are among the largest publicly available cohorts for blood glucose forecasting, making them widely used in current research within this domain~\citep{xie2020glucose_benchmarking, rubin2020deep, rubin_falcone2022glucose_sparse, rabby2021stacked, li2019glucose_glunet}.

\paragraph{Simulated data.} An open-source python implementation of the FDA-approved UVa/Padova Simulator (2008 version) \citep{simglucose_dataset, visentin2016towards_single_day_simulator} provides blood glucose, CHO, and insulin values for 30 simulated patients with Type-I diabetes. We generate 54 days of data for each patient to match the median training set of the OhioT1DM, and approximately 9 days for the test set. Appendix~\ref{apd:first_data} contains examples of blood glucose time series for 6 individuals, shown in Fig.~\ref{fig:simglucose_examples}, as well as basal and bolus dose distributions for 12 individuals, shown in Fig.~\ref{fig:simglucose_insulin_kde}.

\begin{figure}[t!]
\centering
\includegraphics[width=0.4\textwidth]{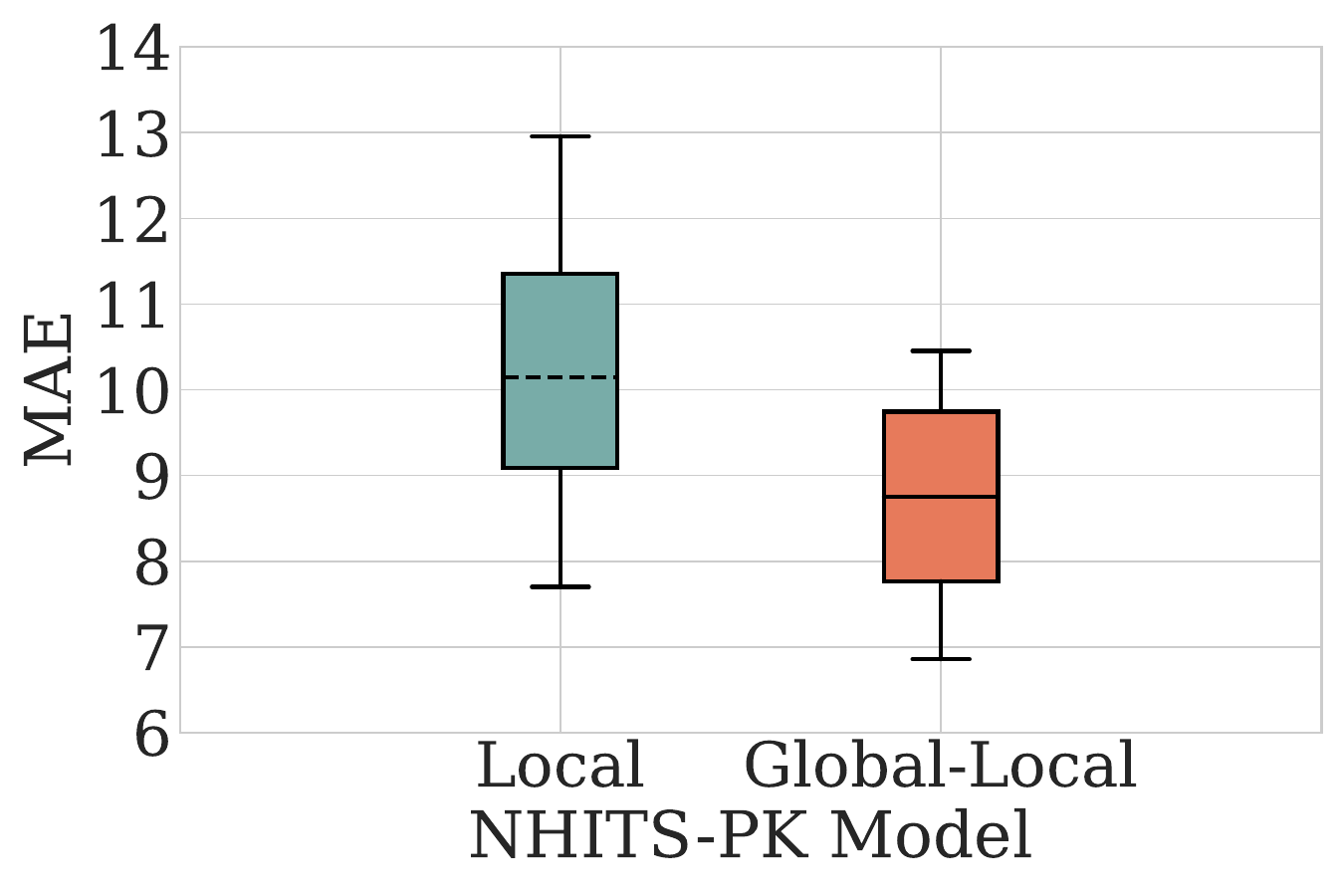}
\caption{The hybrid global-local \NHITS-PK model trained on all OhioT1DM individuals (orange, solid median) has a lower MAE than \NHITS-PK local models (blue, dashed median).}
\label{fig:single_boxplot}
\end{figure}

\begin{figure}[t!]
\centering
\includegraphics[width=0.495\textwidth, trim=13 0 0 0, clip]{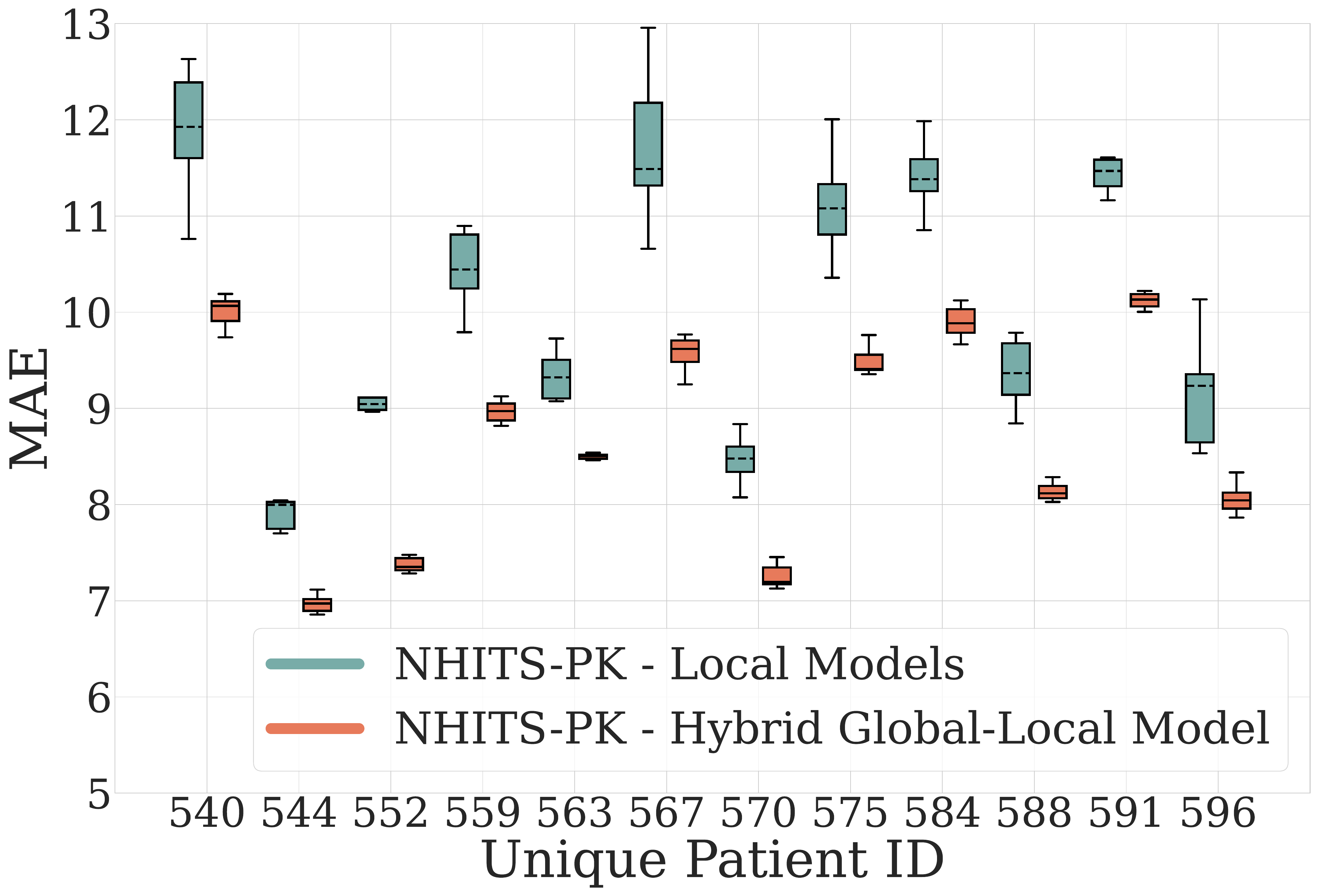}
\caption{The \NHITS-PK hybrid global-local model (orange, solid median) consistently outperforms the local \NHITS-PK model (blue, dashed median).}
\label{fig:ohiot1dm_boxplots}
\end{figure}

\begin{table*}[t!]
\caption{Mean absolute error (MAE) computed for model predictions across all values in the forecast horizon and across only critical values in the forecast horizon (blood glucose $\leq 70~ | \geq 180$). Models are separated with horizontal lines into statistical, linear, RNN-based, CNN-based, Transformer-based, and MLP-based, and PK model groups, respectively. Result for models with (w/) and without (w/o) exogenous (exog.) variables are shown where applicable. Best results are highlighted in \textbf{bold}, second best results are \underline{underlined}. PK models with improved forecasts over their respective exogenous baseline model are highlighted in \textcolor{blue}{blue}.}
\centering
\begin{tabular}{l|cc cc|cc cc}
\toprule
 & \multicolumn{4}{c|}{\textbf{Simulated Dataset}} & \multicolumn{4}{c}{\textbf{OhioT1DM Dataset}} \\
\multirow{3}{*}{\textbf{Model}} & \multicolumn{2}{c}{\textbf{\small{All Values}}} & \multicolumn{2}{c|}{\textbf{\small{Critical Values}}} & \multicolumn{2}{c}{\textbf{\small{All Values}}} & \multicolumn{2}{c}{\textbf{\small{Critical Values}}} \\
{} & \thead{w/o Exog.} & \thead{w/ Exog.} & \thead{w/o Exog.} & \thead{w/ Exog.} & \thead{w/o Exog.} & \thead{w/ Exog.} & \thead{w/o Exog.} & \thead{w/ Exog.} \\
\hline
\ETS & 9.450 & -- & 13.457 & -- & 9.022 & -- & \underline{10.028} & -- \\
\hline
\DLinear & 7.802 & 7.989 & 10.575 & 10.790 & 11.514 & 13.011 & 12.512 & 14.580 \\
\hline
\RNN & 12.728 & 12.930 & 20.308 & 20.945 & 11.535 & 10.907 & 12.641 & 12.989 \\
\LSTM & 9.250 & 9.152 & 12.034 & 12.000 & 10.217 & 10.615 & 12.102 & 12.983 \\
\hline
\TCN & 14.340 & 11.380 & 23.370 & 14.817 & 10.769 & 11.516 & 12.812 & 13.203 \\
\hline
\Informer & 8.599 & -- & 12.074 & -- & 13.714 & -- & 14.867 & -- \\
\PatchTST & 7.284 & -- & 9.842 & -- & 9.922 & -- & 10.999 & -- \\
\TFT & 6.587 & \underline{5.826} & 8.518 & \underline{7.132} & 9.402 & 9.197 & 10.577 & 10.643 \\
\hline
\MLP & 10.424 & 8.407 & 13.932 & 10.297 & 10.300 & 13.254 & 11.550 & 14.997 \\
\NBEATSx & 9.187 & 6.886 & 12.757 & 8.145 & 9.868 & 10.280 & 11.037 & 11.359 \\
\NHITS & 8.268 & 7.304 & 11.114 & 9.094 & 9.060 & \underline{8.873} & 10.337 & 10.137 \\
\hline
\TFT-PK & -- & \textcolor{blue}{\textbf{5.743}} & -- & \textcolor{blue}{\textbf{7.042}} & -- & 9.529 & -- & \textcolor{blue}{10.461} \\
\NBEATSx-PK & -- & \textcolor{blue}{6.806} & -- & \textcolor{blue}{7.998} & -- & \textcolor{blue}{9.983} & -- & \textcolor{blue}{10.927} \\
\NHITS-PK & -- & \textcolor{blue}{7.037} & -- & \textcolor{blue}{8.492} & -- & \textcolor{blue}{\textbf{8.758}} & -- & \textcolor{blue}{\textbf{9.965}} \\
\bottomrule
\end{tabular}
\label{main_results_table_mae}
\end{table*}

\paragraph{Real-world data.} Twelve de-identified individuals with Type 1 diabetes are included in the OhioT1DM 2018 and 2020 datasets \citep{ohiot1dm_dataset}. Each patient has approximately 8 weeks of data that includes blood glucose (mg/dL) measurements recorded with a continuous glucose monitor (CGM) at a frequency of 5-minutes, basal rate information, bolus insulin (units), and CHO values. The maximum number of test samples consistent across all patients (2691 timestamps) was used to obtain equal test sets of approximately 9 days. Missing values were forward filled to prevent data leakage. Sample counts and missing data percentages for each subject are included in Table~\ref{table:real_data_missing_values} in Appendix~\ref{apd:first_data}. Appendix~\ref{apd:first_data} also contains examples of blood glucose time series plots for 6 subjects, shown in Fig.~\ref{fig:ohiot1dm_examples}, as well as basal and bolus dose distributions for the 12 patients, shown in Fig.~\ref{fig:ohiot1dm_insulin_kde}.

\label{section:cohort}
\label{section:results}
\section{Results}
\paragraph{PK-encoder models outperform their sparse exogenous model counterparts.} For simulated data, the \TFT-PK model achieves the lowest MAE, on average, among all models for both evaluations across all models and critical values. The PK models, including \TFT-PK, \NBEATSx-PK, and \NHITS-PK outperform their sparse exogenous model counterparts by up to \TFTSPercMaxBGTG, \NBEATSXSPercMaxBGTG, \NHITSSPercMaxBGTG, respectively, on evaluations across all values and \TFTSPercMaxBGTGc, \NBEATSXSPercMaxBGTGc, \NHITSSPercMaxBGTGc\ on evaluations across critical values for individual patients. For the OhioT1DM dataset, the \NHITS-PK model achieves the lowest MAE, on average, among all models for both evaluations across all values and critical values. The PK models, including, \TFT-PK, \NBEATSx-PK, and \NHITS-PK, outperform their sparse exogenous model counterparts by up to \TFTOPercMaxBGTGc, \NBEATSXOPercMaxBGTGc, \NHITSOPercMaxBGTGc\ on evaluations across critical values for individual patients. Mean MAE and RMSE results and their standard deviations computed across 8 trials are shown in Appendix~\ref{apd:second_mae_rmse} Tables~\ref{mae_table_appendix} and~\ref{rmse_table_appendix}.

\paragraph{Hybrid global-local models significantly improve forecasts over global models.} For the simulated dataset, the \TFT-PK, \NBEATSx-PK, and \NHITS-PK models achieve significantly lower forecast errors for patients for evaluations on all values (p-values: \TFTSpvalBGTG\ [\TFT-PK]; \NBEATSXSpvalBGTG\ [\NBEATSx-PK]; \NHITSSpvalBGTG\ [\NHITS-PK]) and critical values (p-values: \NBEATSXSpvalBGTGc\ [\NBEATSx-PK]; \NHITSSpvalBGTGc\ [\NHITS-PK]) compared to their sparse exogenous global model counterparts. For the OhioT1DM dataset, the \NBEATSx-PK and \NHITS-PK models also achieve significantly lower forecast errors for patients for both evaluations on all values (p-values: \NBEATSXOpvalBGTG\ [\NBEATSx-PK]; \NHITSOpvalBGTG\ [\NHITS-PK]) and critical values (p-values: \NBEATSXOpvalBGTGc\ [\NBEATSx-PK]; \NHITSOpvalBGTGc\ [\NHITS-PK]) compared to their sparse exogenous global model counterparts. The \TFT-PK model does not outperform its sparse exogenous counterpart for $L\!=\!120$. However, for $L\!=\!180$, it achieves significantly lower error (p-value: 4.8e-2) than its sparse exogenous variant and records the lowest error among all \TFT\ model variants, as shown in Table~\ref{tab:contextlen_ablation_table}. Additionally, Table~\ref{table:sumtotal_table_main} shows that \TFT-PK and \NHITS-PK models achieve significantly lower errors, on average—\TFTSPercAvgTGSG\ (p-value: \TFTSpvalTGSG) and \NHITSOPercAvgTGSG\ (p-value: \NHITSOpvalTGSG), respectively—for subjects in evaluations across all values, compared to their ``Sum Total" model counterparts.

\paragraph{Hybrid global-local models significantly improve forecasts over local models.} For the OhioT1DM dataset, the \NHITS-PK hybrid global-local model outperforms local \NHITS-PK models developed for individual OhioT1DM patients by \NHITSOPercAvgTGTL, on average (p-value: \NHITSOpvalTGTL) as shown in Fig.~\ref{fig:single_boxplot}. Furthermore, this improvement in forecast accuracy is consistent across all patients as shown in Fig.~\ref{fig:ohiot1dm_boxplots} and Table~\ref{table:local_global_table_main} (standard deviations are provided in Table~\ref{table:local_global_table} in Appendix~\ref{apd:second_local_results}). Our hybrid global-local \NHITS-PK model also significantly outperforms local sparse exogenous baseline models by \NHITSOPercAvgTGBL, on average, for individual subjects (p-value: \NHITSOpvalTGBL).

\begin{table}[t!]
\caption{PK models outperform Sum Total (ST) models for both Simulated and OhioT1DM datasets. The best results are in \textbf{bold}.}
\centering
\resizebox{0.495\textwidth}{!}{\begin{tabular}{l|cc|cc}
\toprule
{} & \multicolumn{2}{c|}{\textbf{Simulated Dataset}} & \multicolumn{2}{c}{\textbf{OhioT1DM Dataset}} \\
{} & \textbf{\TFT-PK} & \textbf{\TFT-ST} & \textbf{\NHITS-PK} & \textbf{\NHITS-ST} \\
\hline
\multirow{2}{*}{All Values} & \textbf{5.743} & 5.885 & \textbf{8.758} & 8.958 \\
{} & \small{(0.159)} & \small{(0.187)} & \small{(0.080)} & \small{(0.087)} \\
\hline
\multirow{2}{*}{Critical Values} & \textbf{7.042} & 7.233 & \textbf{9.965} & 10.131 \\
{} & \small{(0.270)} & \small{(0.282)} & \small{(0.095)} & \small{(0.100)} \\
\bottomrule
\end{tabular}}
\label{table:sumtotal_table_main}
\end{table}

\begin{table}[t!]
\caption{MAE computed across forecast horizon values for local \NHITS, global \NHITS, local \NHITS-PK, and hybrid global-local \NHITS-PK on the 12 subjects from the OhioT1DM dataset. The best results are in \textbf{bold}.}
\centering
\resizebox{0.385\textwidth}{!}{\begin{tabular}{ r||c c|c c }
 \hline
\multirow{2}{*}{\textbf{ID}} & \multicolumn{2}{c}{\textbf{w/ Exog. (Sparse)}} \vline &  \multicolumn{2}{c}{\textbf{w/ Exog. (PK)}}\\ 
\cline{2-5}
{} & Local & Global & Local & Hybrid\\
 \hline
    540 & 12.106 & 10.119 & 11.880 & \textbf{10.015} \\
    544 & 7.952 & 7.188 & 8.067 & \textbf{6.967} \\
    552 & 8.559 & 7.400 & 9.132 & \textbf{7.374} \\
    559 & 10.845 & 9.013 & 13.051 & \textbf{8.967} \\
    563 & 9.527 & 8.731 & 9.427 & \textbf{8.506} \\
    567 & 12.599 & 9.640 & 12.132 & \textbf{9.562} \\
    570 & 8.361 & 7.523 & 8.466 & \textbf{7.258} \\
    575 & 10.891 & 9.712 & 11.095 & \textbf{9.496} \\
    584 & 11.111 & \textbf{9.846} & 11.518 & 9.905 \\
    588 & 10.172 & 8.172 & 9.376 & \textbf{8.137} \\
    591 & 11.987 & 10.246 & 11.428 & \textbf{10.151} \\
    596 & 8.990 & 8.133 & 9.147 & \textbf{8.059} \\
    \hline
    All & 10.324 & 8.873 & 10.425 & \textbf{8.758} \\
 \hline
\end{tabular}}
\label{table:local_global_table_main}
\end{table}

\begin{figure}[t!]
\centering
\includegraphics[width=0.49\textwidth, trim=15 117 0 0, clip]{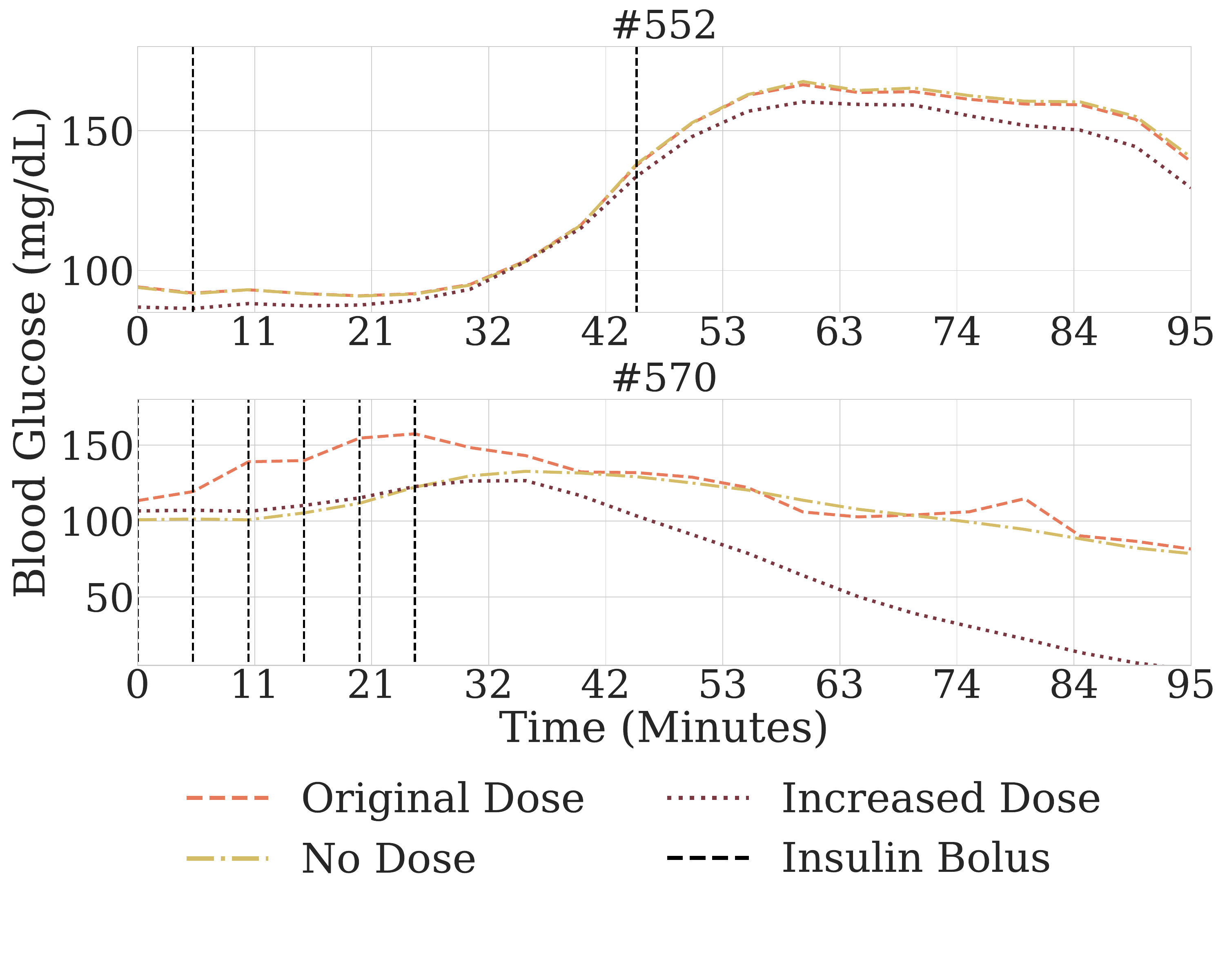}
\caption{The \NHITS-PK model captures insulin treatment effects: forecasts with removed bolus doses (yellow, dash-dotted) and augmented doses (red, dotted), result in higher and lower glucose predictions, respectively, compared to the original doses (orange, dashed). Each point represents the final forecasted time step from a rolling window forecast.}
\label{fig:increased_bolus_dose_forecasts}
\end{figure}

\paragraph{The \NHITS-PK model effectively learns the inhibitory treatment effects of insulin on blood glucose.} We examine whether the \NHITS-PK model can accurately capture treatment effects of insulin on blood glucose by generating forecasts for three scenarios: the original bolus insulin doses, bolus insulin doses removed, and bolus insulin doses increased by a factor of 10. When insulin doses are removed, blood glucose predictions increase, and when insulin doses are augmented, blood glucose predictions decrease as shown in Fig.~\ref{fig:increased_bolus_dose_forecasts}. This result underscores the importance of models that can effectively capture treatment effects, enabling them to adapt to new treatment doses for more reliable and accurate forecasts.

\begin{figure}[t!]
\centering
\includegraphics[width=0.49\textwidth, trim=0 55 0 0, clip]{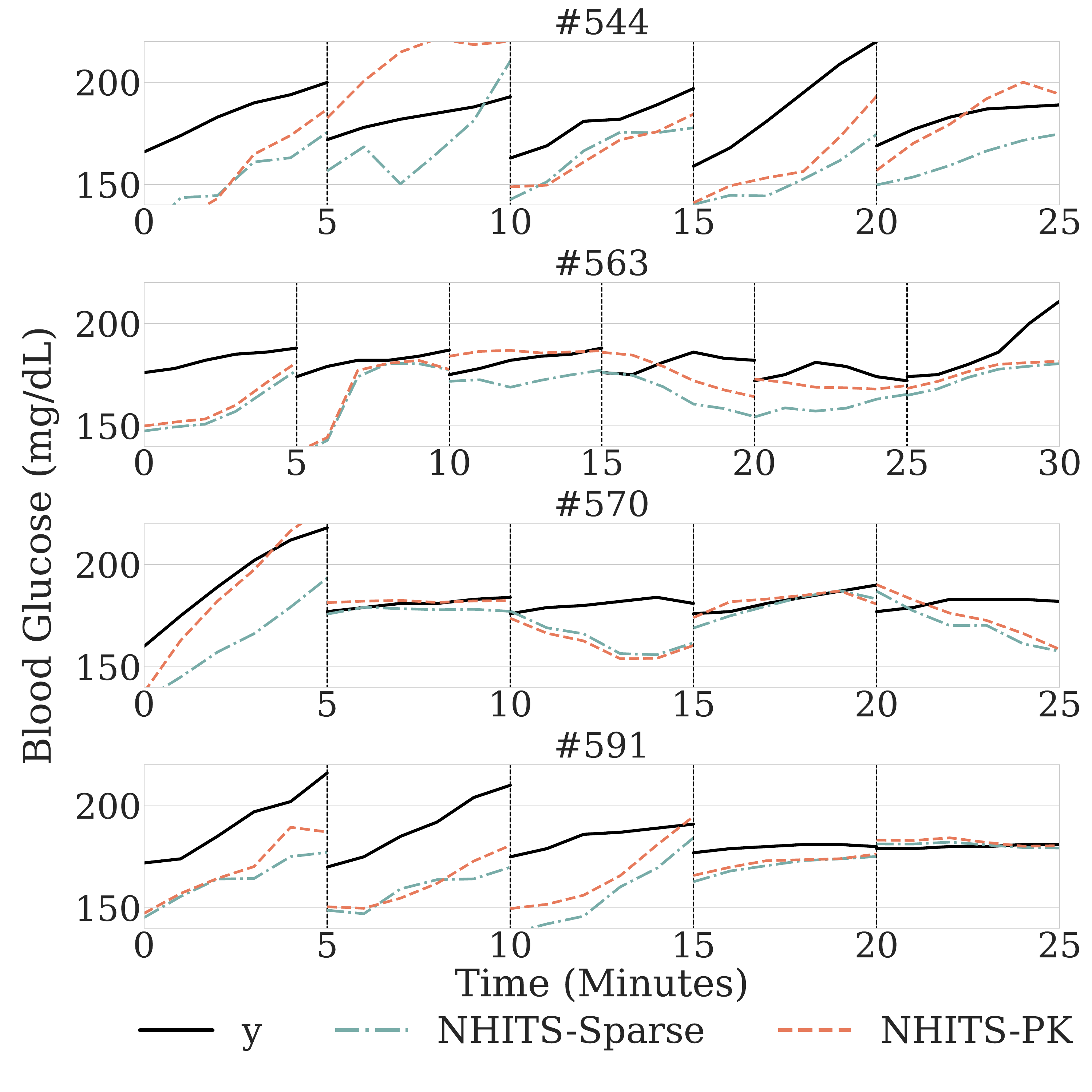}
\caption{Blood glucose forecasts of the final horizon point during 5 randomly selected hyperglycemic periods (blood glucose $\geq$ 180) for 4 randomly selected OhioT1dm patients. Forecasts for the \NHITS-PK model (orange, dashed) are more aligned with ground truth (black, solid) compared to the \NHITS~sparse exogenous model blue, dash-dotted).}
\label{fig:ohiot1dm_forecast_examples}
\end{figure}

\section{Discussion}
\begin{figure}[ht!]
\centering
\includegraphics[width=0.47\textwidth, trim=0 12 0 0, clip]{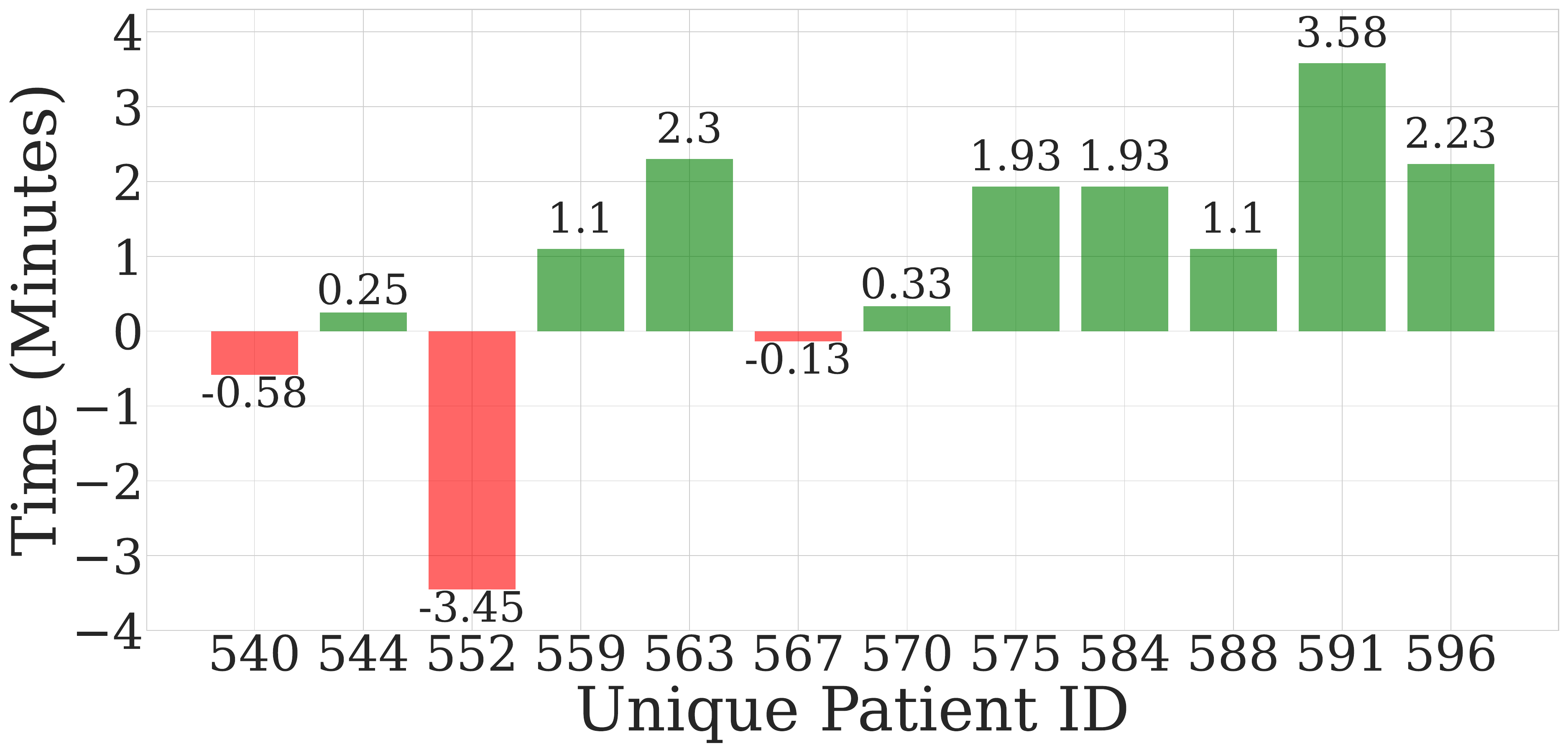}
\caption{The PK encoder enables improved \NHITS\ model prediction lead times for hyperglycemic events (CGM $\geq$ 180) for 9 of 12 OhioT1DM patients.}
\label{fig:timegain_plot}
\end{figure}

The proposed PK encoder can substantially reduce forecasting errors in deep learning models for both simulated and real-world data. In addition to providing more accurate forecasts during hyperglycemic events, our PK encoder predicts these events earlier, up to 3.6 minutes compared to the \NHITS~sparse exogenous baseline model, for individual OhioT1DM patients, as shown in Fig.~\ref{fig:timegain_plot}. The \TFT-PK model achieves a true positive rate of 0.96 on the simulated dataset, while the \NHITS-PK model reaches 0.89 on the OhioT1DM dataset, showcasing their effectiveness in predicting critical events while leaving room for further improvement. A complete analysis of hyper- and hypoglycemic event prediction is provided in Appendix~\ref{apd:tpr_fpr_analysis}.

In addition to improving performance over global model baselines, our hybrid global-local model architecture produces forecasts with significantly smaller errors than patient-specific models. This result is important in demonstrating that global models can integrate cohort-level information with patient-specific parameters to produce more accurate personalized forecasts. Evaluations on the OhioT1DM dataset also demonstrate that our hybrid global-local model remains effective even with 20\% of missing timestamps in the training set for individual patients, as shown in Table~\ref{table:local_global_table_main}, with missing data detailed in Table~\ref{table:real_data_missing_values}. Moreover, our proposed hybrid global-local model offers practical benefits beyond improved forecasting performance as it requires less computational resources and time to train than individual patient models. Training patient-specific \NHITS-PK models took approximately seven times longer than training a single \NHITS-PK hybrid model, despite using a common hyperparameter space. Additionally, the PK models are as computationally efficient as or more efficient than their sparse exogenous counterparts in terms of floating-point operations per second (FLOPs) and the number of trainable parameters, as shown in Table~\ref{tab:efficiency_metrics} in Appendix~\ref{apd:computational_efficiancy_analysis}.

\begin{figure*}[t!]
    \centering
    \begin{minipage}{0.49\textwidth}
        \centering
        \includegraphics[width=\textwidth]{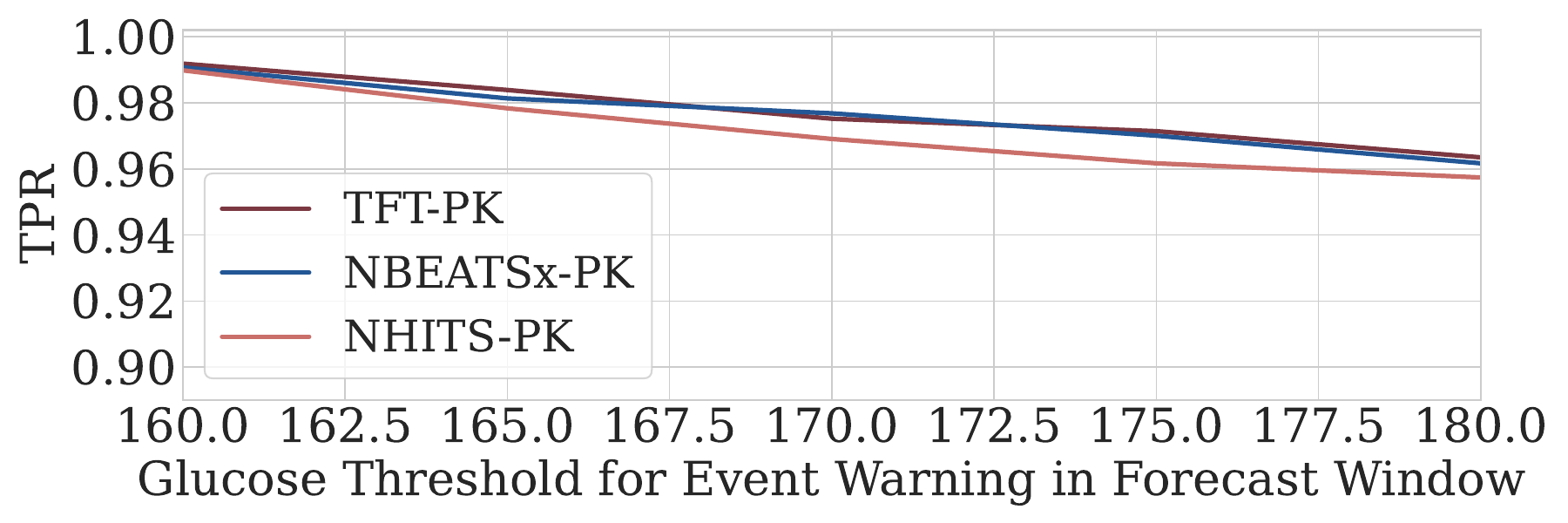}
        (a) Simulated Dataset
    \end{minipage} \hfill
    \begin{minipage}{0.49\textwidth}
        \centering
        \includegraphics[width=\textwidth]{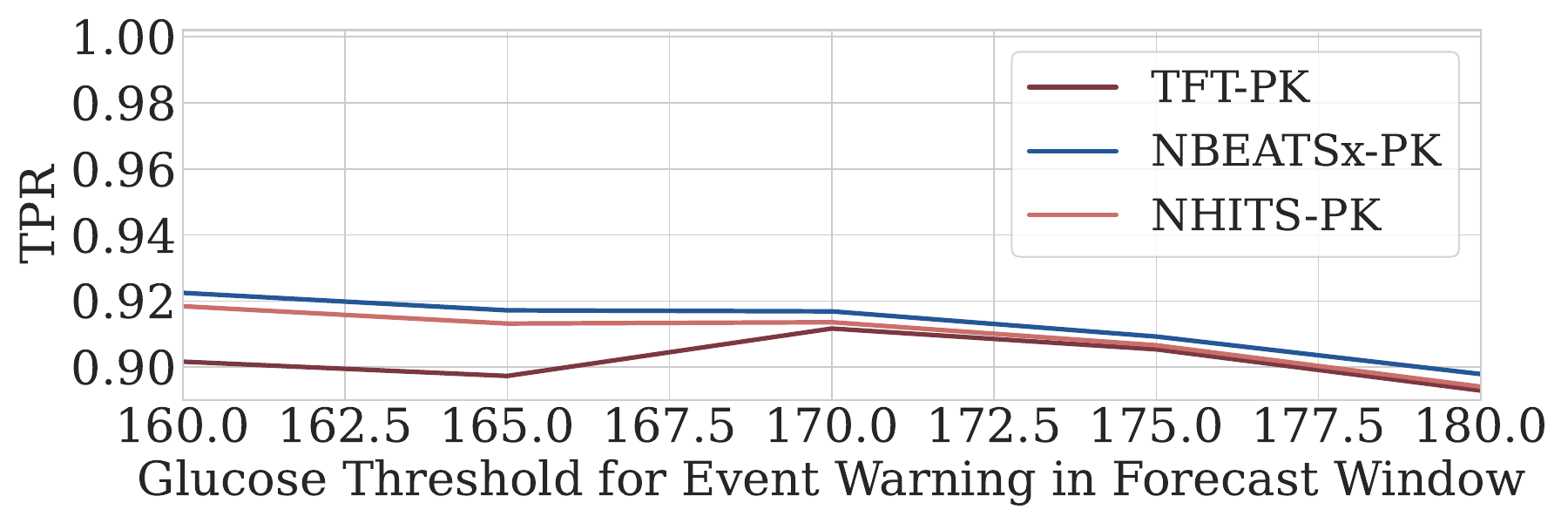}
        (b) OhioT1DM Dataset
    \end{minipage}

    \begin{minipage}{0.49\textwidth}
        \centering
        \includegraphics[width=\textwidth]{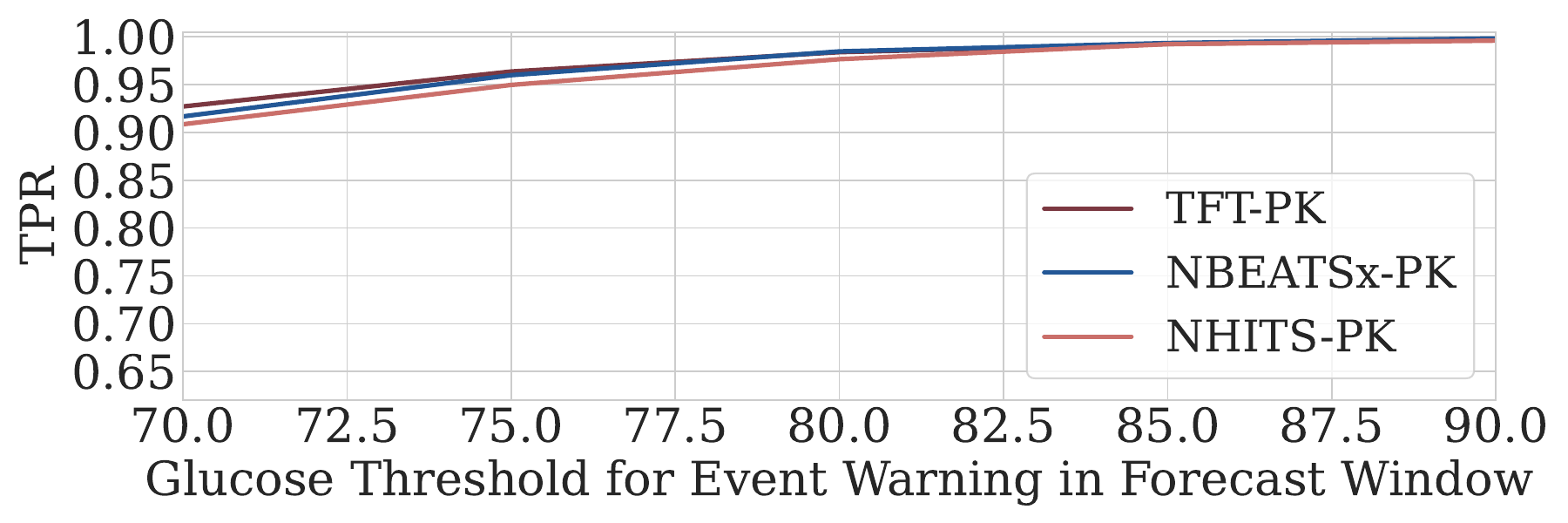}
        (c) Simulated Dataset
    \end{minipage} \hfill
    \begin{minipage}{0.49\textwidth}
        \centering
        \includegraphics[width=\textwidth]{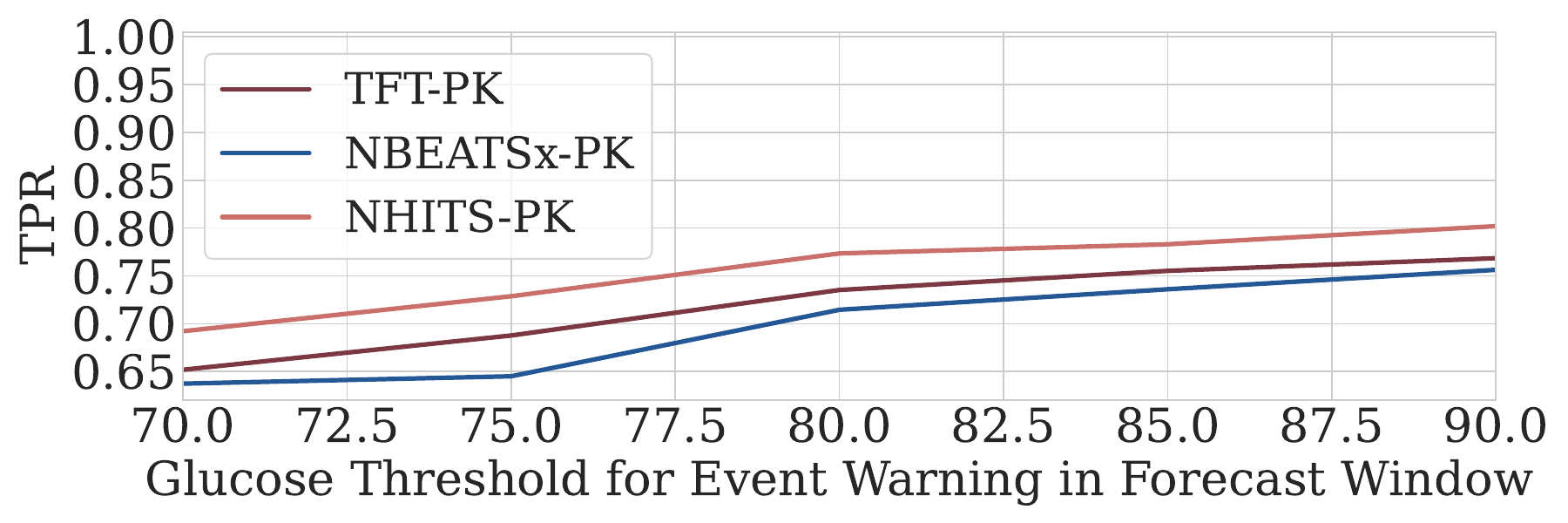}
        (d) OhioT1DM Dataset
    \end{minipage}
    \caption{True positive rate (TPR) for the Simulated and OhioT1DM datasets at various glucose thresholds for \textbf{(a, b) hyperglycemia (blood glucose $\geq 180$)} and \textbf{(c, d) hypoglyemia (blood glucose $\leq 70$)} event warnings. TPR represents the proportion of forecast horizon windows in which the model correctly predicts an event when one actually occurs. The best performing \TFT-PK model for the Simulated dataset predicts hyperglycemic and hypoglycemic events in forecast windows at an approximate TPR of 0.96 and 0.93, respectively. The best performing \NHITS-PK model for the OhioT1DM dataset predicts hyperglycemic and hypoglycemic events in forecast windows at an approximate TPR of 0.89 and 0.69, respectively.}
    \label{fig:hypoglycemia_tpr_fpr_main}
\end{figure*}

Our study demonstrates that leveraging PK domain knowledge to integrate exogenous variables in models offers performance improvements over using sparse features or the ``Sum Total" feature engineering approach. Furthermore, dose-dependent PK encoding strategies, such as those proposed in Section~\ref{section:methods_dose_dependent_effects}, can lead to more realistic models, as highlighted by our ablation study in Appendix~\ref{apd:nonlinear_ablation} where we demonstrate that using our PK encoder to infer the same $\textbf{k}$ value for both bolus and basal insulin, rather than separate values, results in worse performance on the OhioT1DM dataset. 

Our PK encoder can have multiple beneficial applications in clinical practice, such as issuing early warnings of unexpected treatment responses and characterizing patient-specific effects for personalized treatment dosing. We discuss the feasibility of integrating such deep learning models into clinical decision-support systems in section~\ref{apd:clinical_system_feasibility}. While we demonstrate performance improvements across the top three models, our approach can be easily integrated into other deep learning architectures as new tools emerge. It can also be adapted for other types of medication administration, such as oral or intravenous injections.

\paragraph{Limitations and Future Work.} This study has two limitations related to the PK encoder: we assume 100\% bioavailability and a consistent peak concentration across patients. These assumptions focus model learning on the parameter $k$, rather than the area or peak of the concentration curve, to help mitigate potential issues with parameter identification. Future work will explore expanding concentration curve parameters as well as using nonlinear PK frameworks to potentially enable more flexible concentration curves. The proposed method is demonstrated on T1DM patient data but has potential to be adapted to other forecasting tasks and populations, such as insulin-dependent T2DM patients, with appropriate cohort data, PK model adjustments, and model training. Future work will apply our approach to additional cohorts, including T2DM, as more data becomes available to further validate our findings. Future work will also explore methods for estimating patient-specific parameters in a global model for new patients with limited or no historical data.

\label{section:discussion}

\newpage
\acks{This work has been partially supported by the NIH awards R01NR013912 and R01DK131586, and NSF awards 2406231 and 2427948. The authors wish to thank Dr. Frank Guyette, Dr. Leonard Weiss and Dr. Gilles Clermont for their guidance and informative conversations.}

\bibliography{references}

\clearpage
\appendix
\section{Supplemental Methods}
\subsection{Data}\label{apd:first_data}
\paragraph{Simulated data} An open-source python implementation of the FDA-approved UVa/Padova Simulator (2008 version) \citep{simglucose_dataset, visentin2016towards_single_day_simulator} provides clinical and biological parameters for 30 patients (10 adults, 10 adolescents, and 10 children) with Type-I diabetes. Clinical parameters include information on age and weight, and biological parameters include information on insulin-glucose kinetics, such as absorption constants. We generate 54 days of data for each patient to match the median training set of the OhioT1DM, and approximately 9 days for the test set. In a similar approach to \citep{rubin_falcone2022glucose_sparse}, the meal schedule used to generate simulated data was based on the Harrison-Benedict equation \citep{harris_benedict_basal_metabolism} with approximately 3 meals per day and no additional snacks. Height data was not provided in the patient parameters, and so was estimated at 140cm, 170cm, and 175cm for adults, adolescents, and children, respectively.

The `Dexcom' continuous glucose monitor (CGM) option was used to obtain glucose readings at 5-minute intervals. The default basal-bolus controller provided in the simulation was used to administer insulin at the time of each meal, where the bolus amount is computed based on the current glucose level, the target glucose level, the patient's correction factor, and the patient's carbohydrate ratio. However, we adapt the controller to deliver basal doses once per hour, consistent with the function of real insulin pumps used in the OhioT1DM dataset. The resulting simulated data consists of CGM (in mg/dL) values, insulin (units), and carbohydrates (in grams), referred to as CHO. We extract the bolus insulin rate from the overall insulin rate as the difference in the continuous basal insulin rate, which is consistent hourly doses. Blood glucose time series for 6 simulated patients are shown in Fig.~\ref{fig:simglucose_examples} in Appendix \ref{apd:first_data}. Basal and bolus dose distributions for 12 simuluted patients are shown in Fig.~\ref{fig:simglucose_insulin_kde} in Appendix~\ref{apd:first_data}.

\paragraph{OhioT1DM} 12 de-identified individuals with Type 1 diabetes are included in the OhioT1DM 2018 and 2020 datasets \citep{ohiot1dm_dataset}. Each patient has approximately 8 weeks of data that includes blood glucose (mg/dL) measurements recorded with a CGM at a frequency of 5-minutes. The dataset consists of both female (n=5) and male (n=8) patients with unspecified ages within the ranges of 20-40, 40-60, and 60-80. Information on insulin pump model and type of insulin is also provided.

Sparse exogenous features in the data include basal rate (in units per hour), temporary basal rate (units per hour), and bolus insulin (units), and CHO values. Bolus insulin consists of both `normal' insulin, which is recorded as an instantaneous administration, and `square dual', where the insulin dose is stretched over a defined period. `Square dual' insulin doses were divided evenly across the specified administration window with the assumption that insulin units are administered consistently across specified timestamps. If multiple bolus administrations occurred within the same minute, the insulin units were summed. For Medtronic insulin pumps 530G and 630G versions, basal insulin is administered hourly based on the basal rate. For timestamps with overlapping basal rates, the last basal rate value was considered. When specified, the temporary basal rate superseded the basal rate.  

We use the specified train and test partitions provided in the dataset. The maximum number of test samples consistent across all patients (2691 samples) was used to obtain equal test sets of approximately 9 days. The sample count and percent missing data for each subject are included in Table~\ref{table:real_data_missing_values} in Appendix~\ref{apd:first_data}. Missing values were forward filled to prevent data leakage. Blood glucose time series plots for 6 subjects are shown in Fig.~\ref{fig:ohiot1dm_examples} in Appendix~\ref{apd:first_data}. Basal and bolus dose distributions for the 12 patients are shown in Fig.~\ref{fig:ohiot1dm_insulin_kde} in Appendix~\ref{apd:first_data}.

\begin{table}[ht!]
\caption{Sample count (\% missing samples) for the original OhioT1DM datasets and preprocessed test set}
\centering
\begin{tabular}{ r||c|c|c }
 \hline
\multirow{2}{*}{\textbf{ID}} & \multicolumn{2}{c}{\textbf{Original}} \vline &  \multicolumn{1}{c}{\textbf{Preprocessed}}\\ 
\cline{2-4}
{} & Train & Test & Test \\
 \hline
    \multirow{2}{*}{540} & 13109 & 3066  & 2691 \\
    {} & (8.87\%) & (6.43\%) & (6.91\%) \\
     \hline
    \multirow{2}{*}{544} & 12671 & 3137 & 2691  \\
    {} & (16.16\%) &  (13.42\%) & (14.08\%) \\
     \hline
    \multirow{2}{*}{552} & 11097 & 3950 & 2691 \\
    {} & (18.80\%) & (40.15\%) & (53.36\%) \\
     \hline
    \multirow{2}{*}{559} & 12081 & 2876 & 2691 \\
    {} & (10.64\%) & (12.59\%) & (13.45\%) \\
     \hline
    \multirow{2}{*}{563} & 13098 & 2691 & 2691  \\
    {} & (7.44\%) & (4.50\%) & (4.50\%)\\
     \hline
    \multirow{2}{*}{567} & 13536 & 2871 & 2691 \\
    {} & (19.78\%) & (16.79\%) & (16.91\%) \\
     \hline
    \multirow{2}{*}{570} & 11611 & 2880 & 2691  \\
    {} & (5.42\%) & (4.69\%) & (4.91\%) \\
     \hline
    \multirow{2}{*}{575} & 13103 & 2719 & 2691  \\
    {} & (9.44\%) & (4.74\%) & (4.79\%) \\
     \hline
    \multirow{2}{*}{584} & 13248 & 2995 & 2691 \\
    {} & (8.29\%) & (11.02\%) & (11.93\%) \\
     \hline
    \multirow{2}{*}{588} & 13105 & 2881 & 2691 \\
    {} & (3.55\%) & (3.12\%)& (3.34\%) \\
     \hline
    \multirow{2}{*}{591} & 12755 & 2847 & 2691 \\
    {} & (14.96\%) & (3.06\%) & (3.23\%) \\
     \hline
    \multirow{2}{*}{596} & 13630 & 3003 & 2691  \\
    {} & (20.20\%) & (8.66\%)& (9.66\%) \\
 \hline
\end{tabular}
\label{table:real_data_missing_values}
\end{table}

\newpage
\subsection{Models}\label{apd:first_models}

\noindent\textbf{Exponential Smoothing (\ETS)} - The \ETS\ model applies weighted average of past observations and exponentially decreases weights for observations further into the past \citep{ets_2008}. \\

\noindent\textbf{DLinear (\DLinear)} - A linear model that leverages linear layers to model the trend and seasonal components. A moving average filter is used to separate the time series into its trend and seasonal components \citep{zeng_2023_dlinear}. \\

\noindent\textbf{Multi Layer Perceptrons (\MLP)} - A neural network architecture composed of stacked fully connected neural networks trained with backpropagation \citep{nair2010_mlp, fukushima1975_mlp, rosenblatt1958_mlp}. \\

\noindent\textbf{Neural Hierarchical Interpolation for Time Series (\NHITS)} - A deep learning model that applies multi-rate input pooling, hierarchical interpolation, and backcast residual connections together to generate additive predictions with different signal bands. The hierarchical interpolation technique promotes efficient approximations of arbitrarily long horizons and reduced computation \citep{challu_olivares2022_nhits}.\\

\noindent\textbf{Neural Basis Expansion Analysis  with Exogenous (\NBEATSx)} - An MLP-based deep learning model that decomposes the input signal into trend and seasonality components using polynomial and harmonic basis projections. The \NBEATSx\ extension enables modeling of exogenous variables in addition to the univariate signal \citep{oreshkin2020nbeats, OlivaresChallu2022_nbeats}. \\

\noindent\textbf{Recurrent Neural Network (\RNN)} - A recurrent neural network (\RNN) architecture that applies recurrent transformations to obtain hidden states. The hidden states are transformed into contexts which are used as inputs to \MLP s to generate forecasts \citep{elman1990_rnn, cho2014_rnn_encoder_decoder}.\\

\noindent\textbf{Long Short-Term Memory Recurrent Neural Network (\LSTM)} - A recurrent neural network (\RNN) architecture that transforms hidden states from a multilayer \LSTM\ encoder into contexts which are used as inputs to \MLP s to generate forecasts \citep{sak2014_lstm}. \\

\noindent\textbf{Temporal Convolution Network (\TCN)} - A 1D causal-convolutional network architecture that transforms hidden states into contexts which are used as inputs to \MLP\ decoders to generate forecasts. To generate contexts, the prediction at time t is convolved only with elements from time t and earlier in the previous layer in what is referred to as causal convolutions \citep{bai2018_tcn, oord2016_tcn}.\\

\noindent\textbf{Informer (\Informer)} - A Transformer-based deep learning model with a multi-head attention mechanism that uses autoregressive features from a convolution network. A ProbSparse self-attention mechanism and a self-attention distilling process are leveraged to reduce computational complexity \citep{zhou2021informerefficienttransformerlong}. \\

\noindent\textbf{Patch time series Transformer (\PatchTST)} - A Transformer-based deep learning model with a multi-head attention mechanism. The input time series is segmented into fixed-length sub-series-level patches, which serve as input tokens \citep{nie2023patchtst}. \\

\noindent\textbf{Temporal Fusion Transformer (\TFT)} - An attention-based deep learning architecture that learns temporal relationships at different scales using \LSTM s for local processing and self-attention layers to model long-term dependencies. \TFT\ also leverages variable selection networks as well as a series of gating layers to suppress unnecessary parts of the architecture \citep{lim2021_tft}. \\

\subsection{Model Training and Hyperparameters}\label{apd:first_hyperparameters}

The deep learning models were trained using an NVIDIA A100 Tensor Core GPU. For all models, we use the hyperparameter values shown in Table~\ref{tab:hyperpar} and tune the learning rate and random seed, in addition to architecture-specific hyperparameters discussed below. Additionally, for all models with the PK encoder, we tune the initial $\textbf{k}$ values as shown in the second section of table \ref{tab:hyperpar}. We tuned the hyperparameters of our model using `HyperOptSearch' with the Tree-structured Parzen Estimator (TPE) algorithm and 20 samples. 

\begin{table}[htbp]
    \label{tab:hyperparameters}
    \begin{tabular}{l|l} 
      \textbf{Hyperparameter} & \textbf{Considered Values}\\
      \hline
      Learning rate & Loguniform(1e-4, 1e-1) \\
      Batch size & 4 \\
      Windows batch size' & 256 \\
      Dropout & 0.0\\
      Training steps & 2000\\
      Validation check steps & 100 \\
      Early stop & \multirow{2}{*}{5} \\
      patience steps & {} \\
      Random seed & DiscreteRange(1, 10)\\
      \hline
      Initial $\textbf{k}_{\text{basal}}$ & \{1.0, 1.1, 1.2\} \\
      Initial $\textbf{k}_{\text{bolus}}$ & \{1.7, 1.8, 1.9\} \\
      Initial $\textbf{k}_{\text{CHO}}$ & \{1.7, 1.8, 1.9\} \\
      \hline
    \end{tabular}
  \caption{Common hyperparameter search space}
  \label{tab:hyperpar}
\end{table}

In addition to tuning the learning rate and random seed for each model, we also tune architecture-specific hyperparameters. For the \DLinear\ model, we tune the moving average window size ([5, 15, 25, 35]). For the \MLP\ model, we tune the number of hidden units per layer ([128, 256, 512]). For the \LSTM\ and \RNN\ models, we tune the hidden size ([128, 256, 512]). For the \TCN\ model, we tune the hidden size ([128, 256, 512]) and number of dilations ([2, 4, 8]). For the \PatchTST\ model, we tune the hidden size ([128, 256, 512]), number of attention heads ([4, 8, 16]), number of encoder layers ([3, 4]), and patch length ([8, 16, 32, 64, 96, 128]). For the \Informer\ model, we tune the hidden size ([128, 256, 512]), number of attention heads ([4, 8, 16]), number of encoder layers ([3, 4]), and number of decoder layers ([1, 2]). For the \TFT\ model, we tune the hidden size ([128, 256, 512]) and number of attention heads ([4, 8, 16]). For the \NBEATSx\ and \NHITS\ models, we did not tune additional hyperparameters. The code to train models and tune hyperparameters for each architecture is included in the paper's code repository.

\subsection{Model Evaluation}\label{apd:evaluation}

\paragraph{Metrics.} We evaluate forecasts for each test set in the two datasets using two metrics: mean absolute error (MAE) and root mean squared error (RMSE),
\begin{align}
    \mathrm{MAE}(\textbf{y}_{\tau}, \hat{\textbf{y}}_{\tau}) &= \frac{1}{H}\sum_{{\tau}=t+1}^{t+H}|\textbf{y}_{\tau} - \hat{\textbf{y}}_{\tau}| \quad \text{and} \\
    \mathrm{RMSE}(\textbf{y}_{\tau}, \hat{\textbf{y}}_{\tau}) &= \sqrt{\frac{1}{H}\sum_{\tau=t+1}^{t+H}(\textbf{y}_\tau - \hat{\textbf{y}}_\tau)^2}.
\end{align}
Here, $\mathbf{y}_{\tau}$ are the blood glucose values, and $\hat{\mathbf{y}}_{\tau}$ are the model predictions for the forecast horizon $H$, with $\tau = \{T+1, \dots, T+H\}$. We compute the metrics across all points within the forecast horizon and present the average results over 8 trials of independently trained models.

\paragraph{Statistical Significance Tests.} To evaluate the performance differences between our hybrid global-local PK model and the baseline models, we conducted statistical significance testing on the forecasting MAE (Mean Absolute Error) results. For each statistical comparison, patient results were averaged over eight trials of individually trained models, and the paired t-test was computed by pairing patient results between the hybrid global-local PK model and the baseline model. The null hypothesis for each test was that there is no difference in performance between our hybrid global-local PK model and the compared baseline across patients. We considered results statistically significant at the conventional threshold of $\text{p-value} < 0.05$, indicating strong evidence against the null hypothesis.

\subsection{Propositions}\label{apd:first_proposition}

\begin{proposition}
Under linear pharmacokinetic (PK) assumptions, the area under the concentration curve is directly proportional to the dose of the drug. As such, PK parameters, such as bioavailability, are assumed to be constant for a given drug. The bioavailability, $F$, can be assessed from the dose-normalized area under the concentration curve (AUC). When bioavailability is constant, we show that the AUC is bounded by the dose, $x$, given that $0\leq F\leq 1$, 
\label{proposition:auc_bound}
\begin{align}
F &\propto \frac{\text{AUC}}{x}\\
n \cdot F &= n \cdot \frac{\text{AUC}}{x}\\
\text{AUC} &= F\cdot x \\
 &\leq x 
\end{align}

where $n$ where is a constant of proportionality. 
\end{proposition}

\begin{proposition}
Given some mathematical model, $C(t, \cdot)_x$, of concentration for a drug dose, $x$, that is a function of time, $t$, and pharmacokinetic parameters, $\cdot$, we show that the overall effect (concentration) of $n$ drug doses of the same drug is equal to the sum of the individual effects under linear pharmacokinetic assumptions: \label{proposition:concentration_summation}
\begin{align}
    x  &\propto \int_0^T C(t,\cdot)_x dt \\
    n \cdot x &= n\int_0^T C(t,\cdot)_x dt \\
    &= \int_0^T n C(t,\cdot)_x dt \\
    &= \int_0^T \underbrace{C(t,\cdot)_x + C(t,\cdot)_x + \cdots + C(t,\cdot)_x}_{n \text{ times}} dt \\
    &= \int_0^T \sum_{i=1}^n C(t,\cdot)_x dt
\end{align}

Here, $T$ represents the total duration of drug exposure or the time interval over which drug concentrations are being measured and $t$ represents a specific point in time within that interval.
\end{proposition}

\subsection{Algorithm}\label{apd:first_algorithm}
We present the training procedure for the hybrid global-local PK model in Algorithm~\ref{alg:pk_model_training}. This framework is compatible with any deep learning architecture capable of modeling exogenous variables alongside the endogenous target forecast signal.

The algorithm initializes $\textbf{k}$ as a vector of network embedding weights with values between 1 and 2 before model training. The initial value of $\textbf{k}$ is treated as a hyperparameter and tuned using the \texttt{Hyperopt} package from the \texttt{RayTune} library \citep{liaw2018tune}. At each training step, the algorithm iterates over batches of data. The model $f_\theta$ generates forecasts $\hat{\textbf{y}}_\tau$ based on the input data. Network parameters, including $\textbf{k}$, are updated using stochastic gradient descent (SGD) based on the loss of the model $\ell(f_\theta)$, computed between the predicted values $\hat{\textbf{y}}$ and the true values $\textbf{y}$ using the loss function $\mathcal{L}$. After a pre-specified number of training steps, the algorithm performs a validation step by calculating the total validation loss $\ell(f_\theta)$. The final hyperparameters are selected based on the configuration that minimizes validation loss across all runs and are used to train the final model. The algorithm returns the final trained model $f_\theta$.

\begin{algorithm*}
\caption{\texttt{Hybrid Global-Local PK Model Training}. \\$E$ is the number of training steps, $\eta$ is the learning rate, $N$ is the number of subjects, $\nu$ is the sampling interval of the data, $L$ is the lag time, and $H$ is the forecast horizon. The model $f_\theta$ hyperparameters include the learning rate $\eta$, the hidden size $\textbf{h}$, and the initial (init) \textbf{k} values for bolus, basal, and CHO concentration curves, among other hyperparameters. For brevity, we define $\textbf{k}_{\text{init}} = (\textbf{k}_{\text{init, bolus}}, \textbf{k}_{\text{init, basal}}, \textbf{k}_{\text{init, CHO}})$ to represent the initial concentration curve parameters. $\mathcal{L}$ is the loss function (Huber Loss). $V$ is the specified validation step, or the number of training steps between each validation loss check. Forecasts $\hat{\textbf{y}}$ are generated by $f_\theta$ as a function of historical glucose values $\textbf{y}$, treatment concentration curves $\widecheck{\textbf{x}}$, and static features $\textbf{s}$.}
\label{alg:pk_model_training}

\KwIn{$N > 0$, $\nu > 0$, $L>0$, $H>0$, $\eta>0$, $\textbf{h} \in \{128, 256, 512, 1024\}$, $\textbf{k}_{\text{init}} \in [1,2]^{N\times1}$}
\KwOut{$f_\theta$}

\hspace{0.5em}$\mathcal{B}_{\text{train}} \leftarrow$ \textnormal{(split training data into batches)}\\
$\mathcal{B}_{\text{val}} \leftarrow$ \textnormal{(split validation data into batches)}\\

\For{\textnormal{each run $o$ in hyperparameter tuning samples (runs) $O$}}{
    \hspace{0.5em}$f_\theta \leftarrow$ \textnormal{(initialize the model)}\\
    $\textbf{k} \leftarrow \textbf{k}_{\text{init}}$ \Comment{Each entry in $\textbf{k}$ is initialized to the same scalar value} \\
    $\ell(f_\theta)_{\text{val}}^{(o)} \leftarrow 0$ \Comment{Initialize validation loss for run $o$}\\

    \For{\textnormal{each step $i$ from $1$ to $E$}}{
        \For{batch $b \in \mathcal{B}_{\text{train}}$}{
            \hspace{0.5em}$\widecheck{\textbf{x}} \leftarrow \sum_{t=0}^{L} C(\nu\textbf{w}_{t, :}, \textbf{x}_t, \textbf{k})$ \\[1ex]
            $\hat{\textbf{y}}_\tau \leftarrow f_\theta(\textbf{y}, \widecheck{\textbf{x}}, \textbf{s})$\\[1ex]
            $\ell_{\text{batch}}(f_\theta) \leftarrow \mathcal{L}(\textbf{y}_\tau, \hat{\textbf{y}}_\tau)$ \Comment{$\tau$ represents the forecast horizon} \\[1ex]
            $\theta \leftarrow \theta - \eta \cdot \nabla_{\theta} \ell_{\text{batch}}(f_\theta)$ \Comment{update network parameters using SGD}
        }

        \If{$i \mod V == 0$}{
            \hspace{0.5em}$\ell(f_\theta)_{\text{total}} \leftarrow 0$\\
            \For{batch $b \in \mathcal{B}_{\text{val}}$}{
                \hspace{0.5em}$\widecheck{\textbf{x}} \leftarrow \sum_{t=0}^{L} C(\nu\textbf{w}_{t, :}, \textbf{x}_t, \textbf{k})$ \\[1ex]
                $\hat{\textbf{y}}_\tau \leftarrow f_\theta(\textbf{y}, \widecheck{\textbf{x}}, \textbf{s})$\\[1ex]
                $\ell_{\text{batch}}(f_\theta)  \leftarrow \mathcal{L}(\textbf{y}_\tau, \hat{\textbf{y}}_\tau)$ \\[1ex]
                $\ell(f_\theta)_{\text{total}} \leftarrow \ell(f_\theta)_{\text{total}} + \ell_{\text{batch}}(f_\theta) $
            }
            $\ell(f_\theta)_{\text{val}}^{(o)} \leftarrow \frac{\ell(f_\theta)_{\text{total}}}{|\mathcal{B}_{\text{val}}|}$ \Comment{Average loss over entire validation set and update loss}
        }
    }
}

\hspace{0.5em}$\{\eta, \textbf{h}, \textbf{k}_{\text{init}}, \ldots\} \leftarrow \arg \min_{o \in O} \ell(f_\theta)_{\text{val}}$ \Comment{select best hyperparameters; retrain final model} \\
\textbf{return} $f_\theta$
\end{algorithm*}

\newpage
\begin{figure*}[ht!]
\centering
\includegraphics[width=1.0\textwidth]{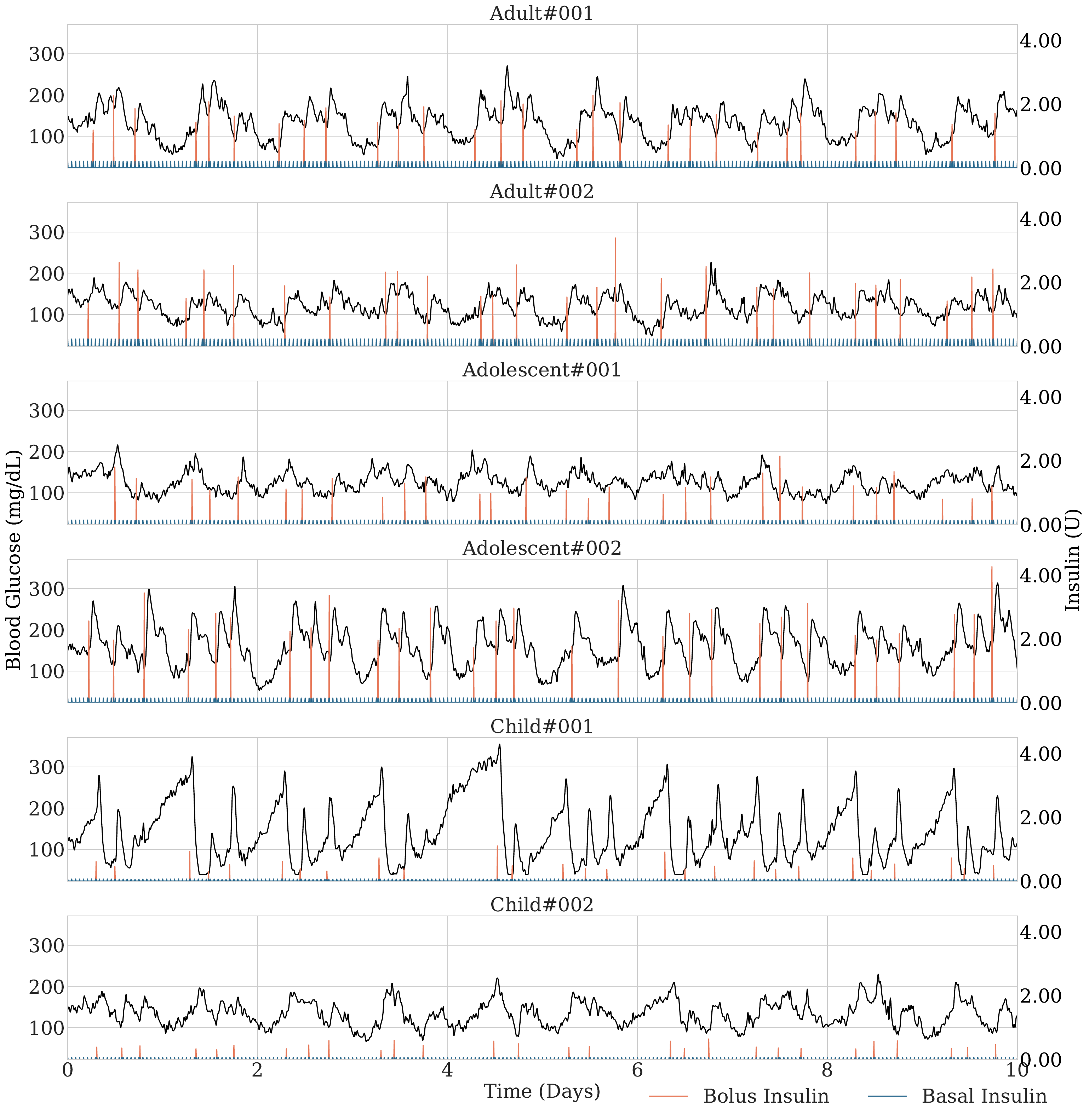}
\caption{Time series for the first 10 days of 6 patients from the simulated dataset.}
\label{fig:simglucose_examples}
\end{figure*}

\clearpage
\begin{figure*}[ht!]
\centering
\includegraphics[width=1.0\textwidth]{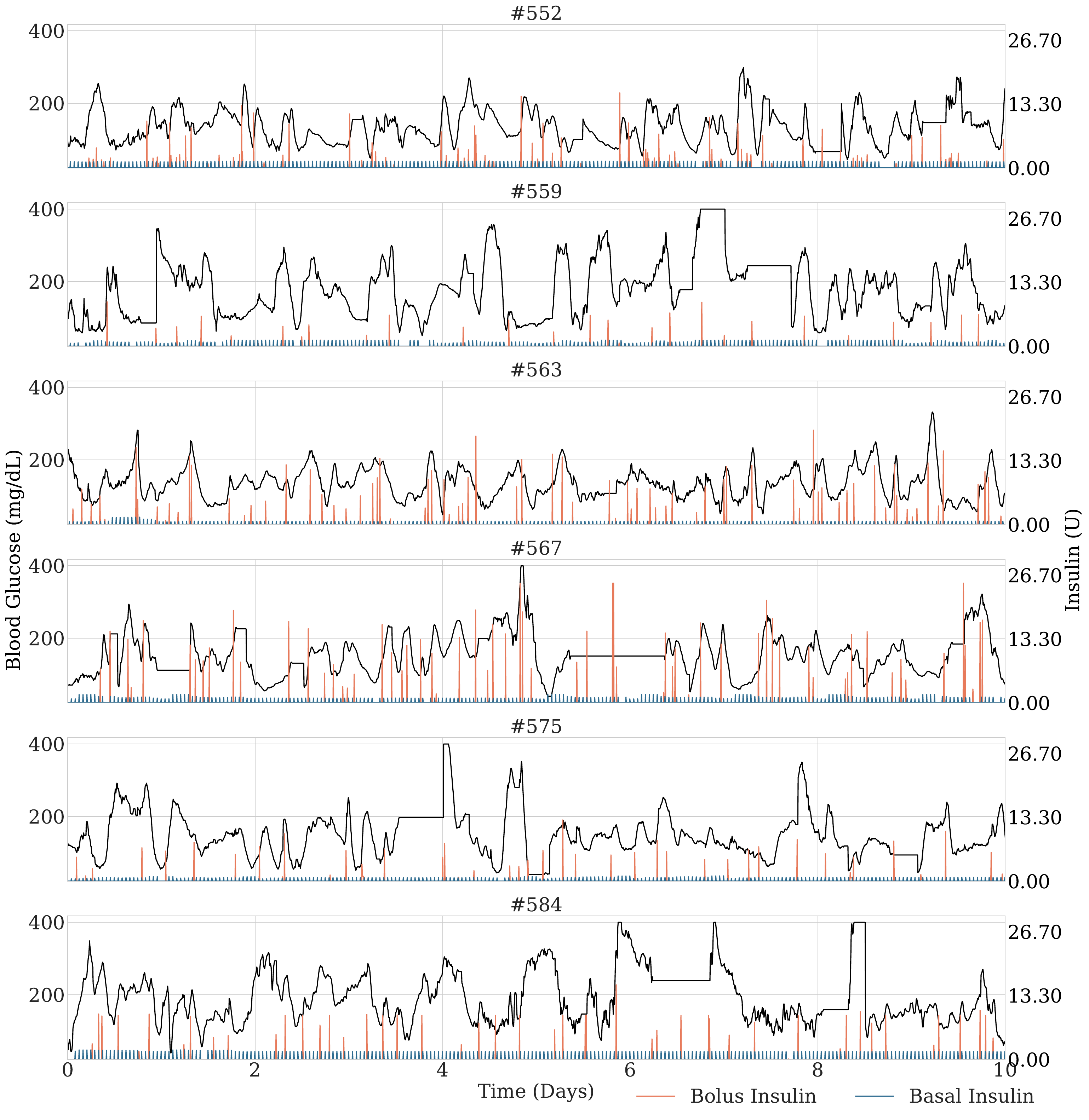}
\caption{Time series for the first 10 days of 6 patients from the OhioT1DM dataset.}
\label{fig:ohiot1dm_examples}
\end{figure*}

\clearpage
\begin{figure*}[ht!]
\centering
\includegraphics[width=1.0\textwidth]{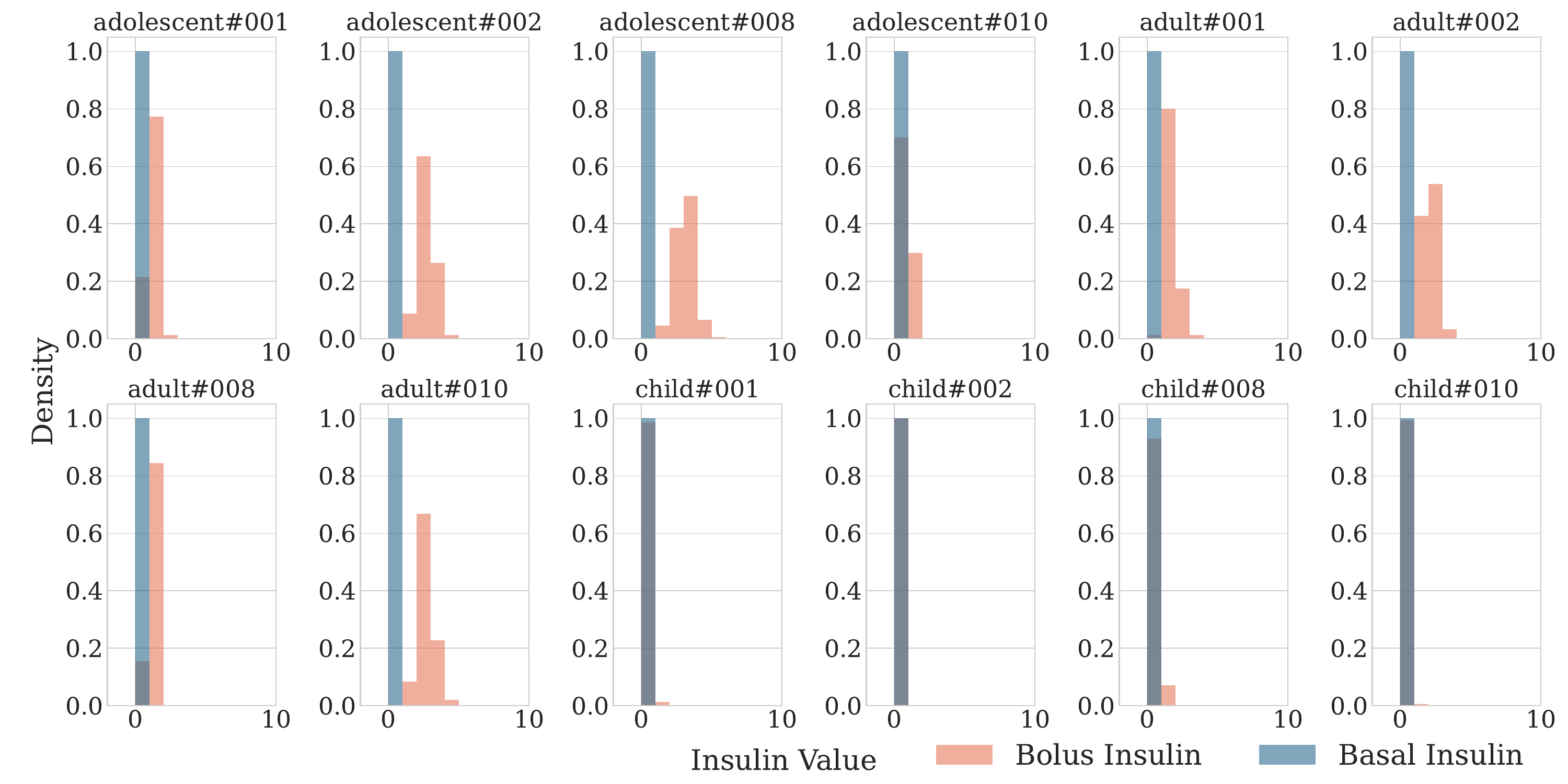}
\caption{Density plots of bolus and basal insulin for 12 simulated data patients. Bolus doses are general larger than basal doses.}
\label{fig:simglucose_insulin_kde}
\end{figure*}

\begin{figure*}[ht!]
\centering
\includegraphics[width=1.0\textwidth]{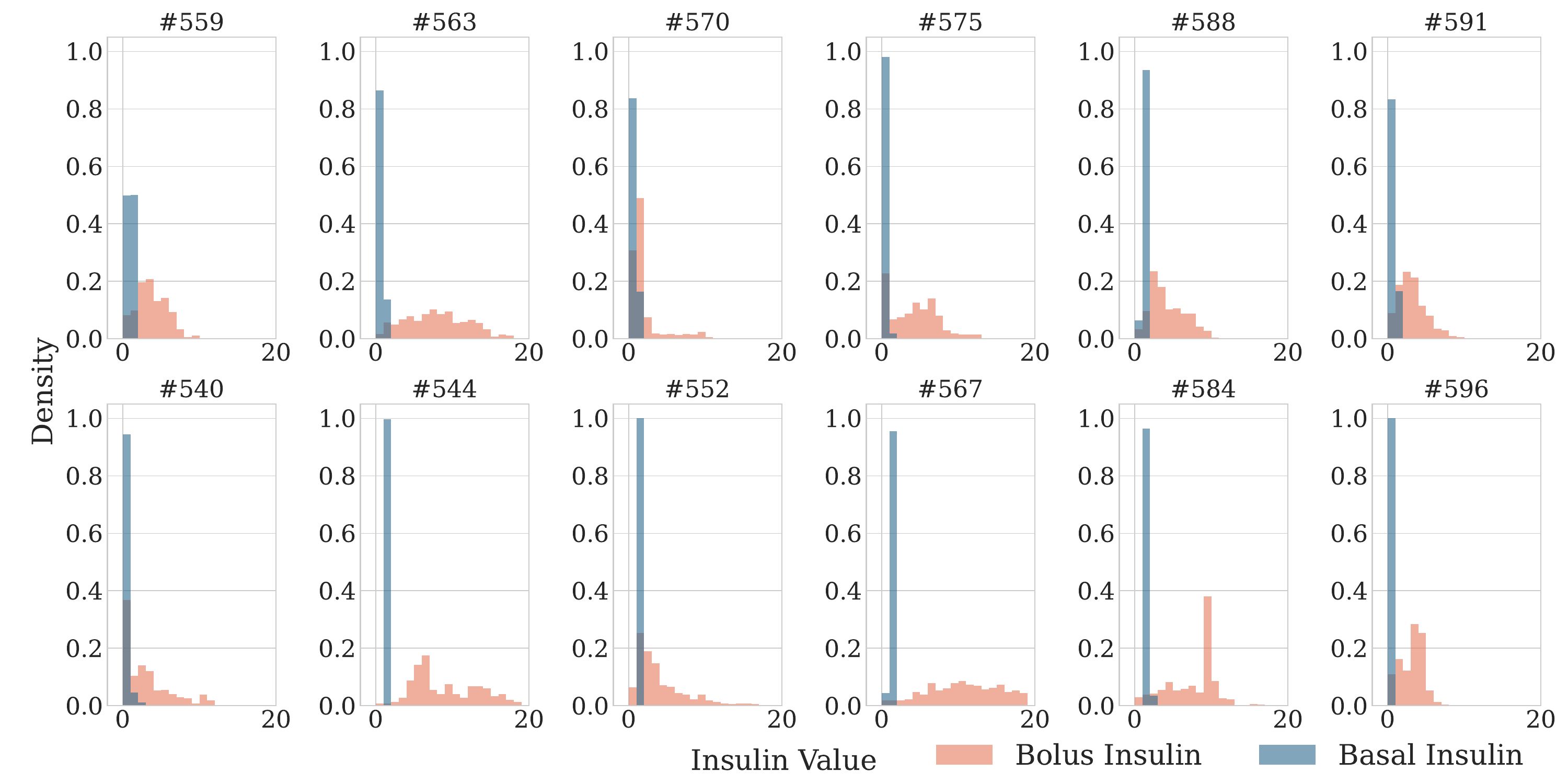}
\caption{Density plots of bolus and basal insulin for OhioT1DM patients. Bolus doses are general larger than basal doses.}
\label{fig:ohiot1dm_insulin_kde}
\end{figure*}

\clearpage

\newpage
\section{Pharmacokinetic Background}\label{apd:pk_background}

\paragraph{Concentration-time Profiles} From PK literature, exponential decay functions are commonly used to model drug concentration over time for intravenous injections, where the drug is directly injected into the bloodstream, leading to instantaneous effects \citep{basic_pharmacokinetics, insulin2005analogs}. Alternatively, subcutaneous injections involve absorption and elimination processes where the peak onset time may not be instantaneous depending on the drug profile. Regarding knowledge on concentration-time profiles established through empirical and invasive studies, we know that concentration-time profiles for subcutaneous injections adhere to right-skewed curves. The degree of right-skew in the concentration curve will differ depending on the type of insulin, as illustrated in Fig.~\ref{fig:bolus_basal_diagram}, along with various patient-specific factors. 

\paragraph{Insulin Analogs} Insulin analogs are modified forms of insulin designed to exhibit certain PK properties, such as changes in drug onset and duration of action. Fig.~\ref{fig:bolus_basal_diagram} is adapted from PK literature and depicts PK modeling of plasma insulin concentration over time for different insulin analogs \citep{insulin2005analogs}. Generally, bolus doses are quick-acting and, thus, often taken before meals. Contrastingly, basal insulin is longer-acting and serves to regulate blood glucose levels throughout the day. However, continuous glucose monitors (CGMs) and insulin pumps generally only hold and administer one type of insulin, such as rapid-acting insulin. For these devices, `bolus' and `basal' are terms used to refer to different timings of injections for different purposes. Bolus doses are administered discretely, generally around meals, while basal doses are administered continuously at a defined frequency, such as hourly, to regulate blood glucose levels throughout the day.

\paragraph{Bioavailability} In Fig.~\ref{fig:bolus_basal_diagram}, the area under the concentration curve measures drug \emph{bioavailability}, the fraction of the absorbed dose that reaches the intended site of systemic circulation intact. Bioavailability varies based on many factors, including type of medication, dosage, and patient-specific characteristics. For instance, the bioavailability of an intravenous dose of any drug is close to 100 percent~\citep{soeborg2012bioavailability}. However, for subcutaneous injections, bioavailability is typically lower due to unintended drug absorption or elimination.

\paragraph{First-order Kinetics} First-order kinetics, a common assumption in pharmacokinetics, describes the rate at which a drug is eliminated from the body as being proportional to its concentration. This means that a constant fraction of the drug is eliminated per unit of time, regardless of its concentration levels. Exponential functions are often used to represent concentration-time profiles under the assumption of first-order kinetics, simplifying the modeling of plasma drug concentration in the body.

\paragraph{Linear Pharmacokinetics} 
Linear pharmacokinetics refers to a situation where the relationship between the dose of a drug and its concentration in the body is linear \citep{basic_pharmacokinetics}. In linear pharmacokinetics, the area under the concentration curve is directly proportional to the dose of the drug. If you double the dose of a drug, the area under the curve will also double. Linearity results from the assumption that the drug processes exhibit first-order kinetics and that the PK parameters remain constant with dose, indicating no significant changes in the body's ability to eliminate the drug with higher doses. The majority of drugs used in clinical practice are assumed to follow linear pharmacokinetics \citep{basic_pharmacokinetics}.

\begin{figure}[H]
    \centering
    \includegraphics[width=0.5\textwidth]{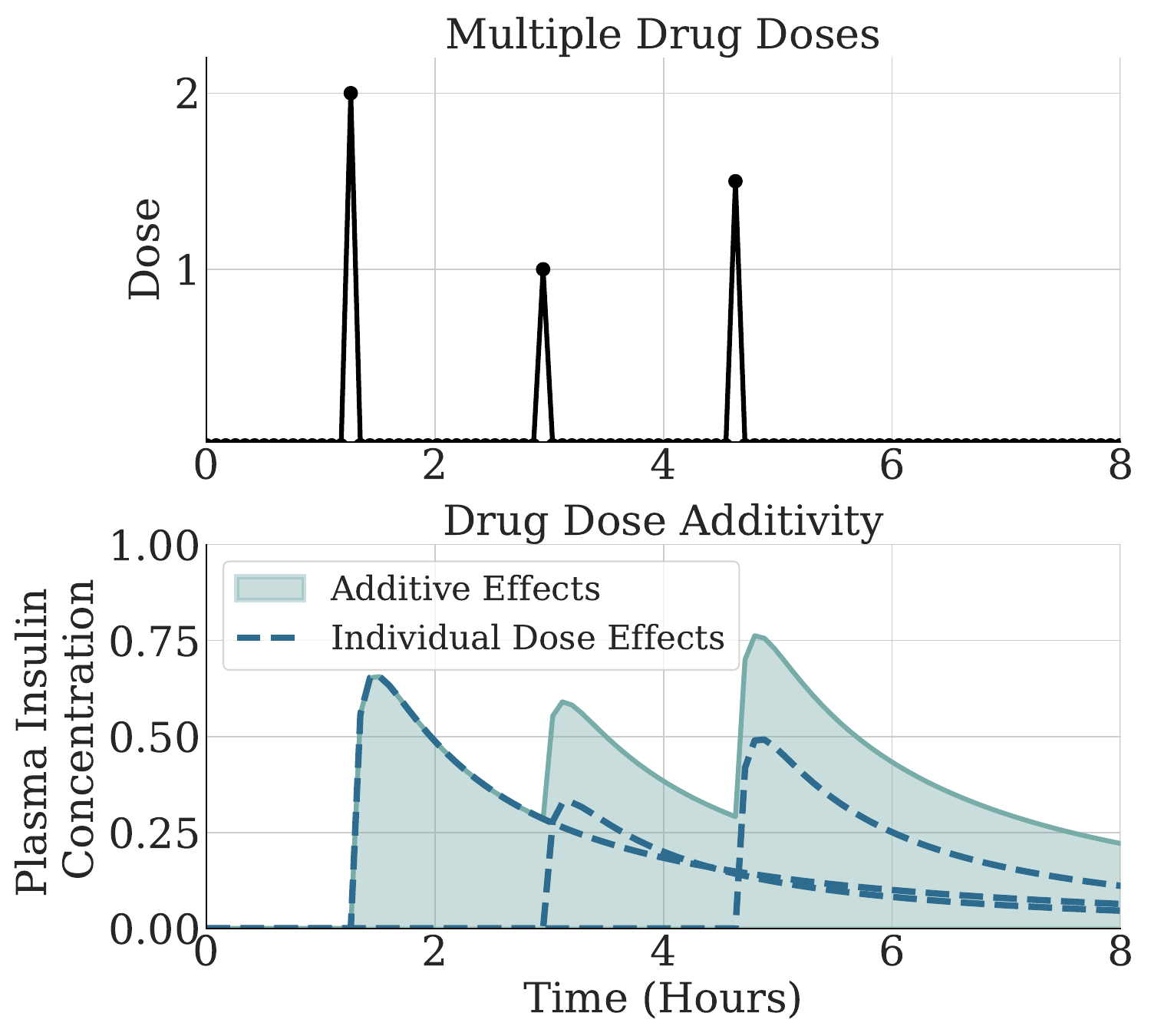}
    \caption{When doses of insulin are administered at close intervals, the subsequent dose may start to act before the previous dose has fully finished its effect. This can lead to a stacking of insulin effects. The figure depicts additive effects under linear pharmacokinetic assumptions.}
    \label{fig:insulin_stacking}
\end{figure}  

\newpage
\section{Additional Results}

\subsection{Patient-specific (Local) Models}\label{apd:second_local_results}
We showcase the effectiveness of global models, with parameters shared across subjects, in generating more accuracy forecasts compared to local models trained for individual subjects. Global PK models consistently outperform local PK models, on average across 8 trials, for each OhioT1DM patient as shown in Table~\ref{table:local_global_table} and Fig.~\ref{fig:ohiot1dm_boxplots}. Furthermore, the global \NHITS-PK model outperforms the global \NHITS sparse exogenous model for 11 out of 12 OhioT1DM patients, on average across 8 trials, as shown in Table~\ref{table:local_global_table}.

\begin{table}[ht]
\caption{MAE computed across forecast horizon values using local and global models for the OhioT1DM dataset. The best results are in \textbf{bold}.}
\centering
\begin{tabular}{ r||c c|c c }
 \hline
\multirow{2}{*}{\textbf{ID}} & \multicolumn{2}{c}{\textbf{w/ Exog. (Sparse)}} \vline &  \multicolumn{2}{c}{\textbf{w/ Exog. (PK)}}\\ 
\cline{2-5}
{} & Local & Global & Local & Global\\
 \hline
    \multirow{2}{*}{540} & 12.106 & 10.119 & 11.880 & \textbf{10.015} \\
    {} & \small{(0.779)} & \small{(0.139)} & \small{(0.596)} & \small{(0.147)} \\
    \multirow{2}{*}{544} & 7.952 & 7.188 & 8.067 & \textbf{6.967} \\
    {} & \small{(0.679)} & \small{(0.062)} & \small{(0.490)} & \small{(0.083)} \\
    \multirow{2}{*}{552} & 8.559 & 7.400 & 9.132 & \textbf{7.374} \\
    {} & \small{(0.243)} & \small{(0.118)} & \small{(0.403)} & \small{(0.070)} \\
    \multirow{2}{*}{559} & 10.845 & 9.013 & 13.051 & \textbf{8.967} \\
    {} & \small{(1.266)} & \small{(0.176)} & \small{(7.027)} & \small{(0.102)} \\
    \multirow{2}{*}{563} & 9.527 & 8.731 & 9.427 & \textbf{8.506} \\
    {} & \small{(0.379)} & \small{(0.065)} & \small{(0.402)} & \small{(0.084)} \\
    \multirow{2}{*}{567} & 12.599 & 9.640 & 12.132 & \textbf{9.562} \\
    {} & \small{(1.649)} & \small{(0.138)} & \small{(1.667)} & \small{(0.189)} \\
    \multirow{2}{*}{570} & 8.361 & 7.523 & 8.466 & \textbf{7.258} \\
    {} & \small{(0.496)} & \small{(0.141)} & \small{(0.223)} & \small{(0.121)} \\
    \multirow{2}{*}{575} & 10.891 & 9.712 & 11.095 & \textbf{9.496} \\
    {} & \small{(0.419)} & \small{(0.099)} & \small{(0.470)} & \small{(0.151)} \\
    \multirow{2}{*}{584} & 11.111 & \textbf{9.846} & 11.518 & 9.905 \\
    {} & \small{(0.547)} & \small{(0.091)} & \small{(0.538)} & \small{(0.156)} \\
    \multirow{2}{*}{588} & 10.172 & 8.172 & 9.376 & \textbf{8.137} \\
    {} & \small{(1.737)} & \small{(0.081)} & \small{(0.321)} & \small{(0.087)} \\
    \multirow{2}{*}{591} & 11.987 & 10.246 & 11.428 & \textbf{10.151} \\
    {} & \small{(1.274)} & \small{(0.083)} & \small{(0.172)} & \small{(0.134)} \\
    \multirow{2}{*}{596} & 8.990 & 8.133 & 9.147 & \textbf{8.059} \\
    {} & \small{(0.246)} & \small{(0.103)} & \small{(0.502)} & \small{(0.141)} \\
    \hline
    \multirow{2}{*}{All} & 10.324 & 8.873 & 10.425 & \textbf{8.758} \\
    {} & \small{(0.401)} & \small{(0.064)} & \small{(0.659)} & \small{(0.080)} \\
 \hline
\end{tabular}
\label{table:local_global_table}
\end{table}

\subsection{Forecasting with Dose-Independent Parameters}\label{apd:nonlinear_ablation}

\begin{table}[ht!]
\caption{Mean absolute error (MAE) computed for model predictions across all values in the forecast horizon and across only critical values in the forecast horizon (blood glucose $\leq 70 | \geq 180$). Result for models with (w/) and without (w/o) exogenous variables are shown. The best results are in \textbf{bold}.}
\centering
\resizebox{0.49\textwidth}{!}{\begin{tabular}{l||cc|cc}
\hline
\multicolumn{5}{c}{\textbf{Simulated Dataset}}\\
\hline
\multirow{2}{*}{\textbf{Model}} &  \multicolumn{2}{c}{\textbf{All Values}} & \multicolumn{2}{c}{\textbf{Critical Values}}\\
\cline{2-5}
{} & \thead{w/o Exog.} & \thead{w/ Exog.} & \thead{w/o Exog.} & \thead{w/ Exog.}\\
\hline
\multirow{2}{*}{\NHITS} & 8.268 & 7.304 & 11.114 & 9.094 \\
{} & \small{(0.039)} & \small{(0.055)} & \small{(0.237)} & \small{(0.236)} \\
\multirow{2}{*}{\NHITS-ST} & \multirow{2}{*}{--} & 7.055 & \multirow{2}{*}{--} & 8.648 \\
{} & {} & \small{(0.038)} & {} & \small{(0.068)} \\
\hline
\multirow{2}{*}{\NHITS-PK} & \multirow{2}{*}{--} & 7.103 & \multirow{2}{*}{--} & \textbf{8.449}  \\
\small{linear} & {} & \small{(0.069)} & {} & \small{(0.139)} \\
\multirow{2}{*}{\NHITS-PK} & \multirow{2}{*}{--} & \textbf{7.037} & \multirow{2}{*}{--} & 8.492 \\
\small{nonlinear} & {} & \small{(0.093)} & {} & \small{(0.220)} \\
\hline
\hline

\hline
\multicolumn{5}{c}{\textbf{OhioT1DM Dataset}}\\
\hline
\multirow{2}{*}{\NHITS} & 9.060 & 8.873 & 10.337 & 10.137 \\
{} & \small{(0.142)} & \small{(0.064)} & \small{(0.137)} & \small{(0.108)} \\
\multirow{2}{*}{\NHITS-ST} & \multirow{2}{*}{--} & 8.958 & \multirow{2}{*}{--} & 10.131 \\
{} & {} & \small{(0.087	)} & {} & \small{(0.100)} \\
\hline
\multirow{2}{*}{\NHITS-PK} & \multirow{2}{*}{--} & 8.830 & \multirow{2}{*}{--} & 10.026  \\
\small{linear} & {} & \small{(0.060)} & {} & \small{(0.064)} \\
\multirow{2}{*}{\NHITS-PK} & \multirow{2}{*}{--} & \textbf{8.758} & \multirow{2}{*}{--} & \textbf{9.965}  \\
\small{nonlinear} & {} & \small{(0.080)} & {} & \small{(0.089)} \\
\hline
\end{tabular}}
\label{ablation_results_table}
\end{table}

We conducted an ablation study to assess whether modeling patient-specific effects of bolus and basal insulin under the assumption of dose-independent effects results in improved forecasting performance. We use our PK encoder to infer the same $\textbf{k}$ for both bolus and basal insulin instead of
separate values as shown in Table~\ref{ablation_results_table} for \NHITS-PK$_{\text{linear}}$. We compare these results to \NHITS-PK$_{\text{nonlinear}}$, where the PK encoder infers separate $\textbf{k}$ values for bolus and basal insulin. A separate $\textbf{k}$ value is inferred for the CHO variable, consistent with the method in the main paper.

We found that learning a single $\textbf{k}$ value for both bolus and basal insulin improved performance on the simulated dataset only when forecasts were evaluated at critical values. However, this approach resulted in worse performance when evaluated across all values in the simulated dataset, as well as on the OhioT1DM dataset. The improved performance on the simulated dataset may be due to the simulated data generator relying on linear PK by treating bolus and basal doses in aggregate and using static parameters for insulin-glucose kinetics \citep{simglucose_dataset, visentin2016towards_single_day_simulator}. In contrast, insulin may exhibit nonlinear dose-dependent PK in real-world scenarios, as mentioned in section \ref{section:methods_dose_dependent_effects}.

\begin{table*}[t!]
\caption{Comparing lag time $L$ effect on model performance in terms of mean absolute error (MAE) computed for model predictions across all values in the forecast horizon and only across critical values in the forecast horizon (blood glucose $\leq 70 | \geq 180$) for simulated dataset and OhioT1DM dataset. Average results are computed across 8 trials. Best results are highlighted in \textbf{bold}.}
\centering
\resizebox{1.0\textwidth}{!}{\begin{tabular}{l|l| cc cc|cc cc}
\toprule
{} & {} & \multicolumn{4}{c|}{\textbf{Simulated Dataset}} & \multicolumn{4}{c}{\textbf{OhioT1DM Dataset}} \\

\multirow{2}{*}{\textbf{Lag Time ($L$)}}& \multirow{2}{*}{\textbf{Model}} & \multicolumn{2}{c}{\textbf{\small{All Values}}} & \multicolumn{2}{c|}{\textbf{\small{Critical Values}}} & \multicolumn{2}{c}{\textbf{\small{All Values}}} & \multicolumn{2}{c}{\textbf{\small{Critical Values}}} \\

\textbf{} & {} & \thead{\textbf{w/ Exog.}} & \thead{\textbf{w/ Exog. (PK)}} & \thead{\textbf{w/ Exog.}} & \thead{\textbf{w/ Exog. (PK)}} & \thead{\textbf{w/ Exog.}} & \thead{\textbf{w/ Exog. (PK)}} & \thead{\textbf{w/ Exog.}} & \thead{\textbf{w/ Exog. (PK)}} \\
\hline

\multirow{5}{*}{$L=30$ timesteps} & \multirow{2}{*}{\TFT} & \textbf{5.836} & 6.064 & \textbf{7.318} & 7.756 & \textbf{9.136} & 9.145 & \textbf{10.271} & 10.464 \\
{} & {} & \small{(0.248)} & \small{(0.466)} & \small{(0.364)} & \small{(0.847)} & \small{(0.363)} & \small{(0.480)} & \small{(0.260)} & \small{(0.790)} \\

\multirow{3}{*}{(2.5 hours)} & \multirow{2}{*}{\NBEATSx} & 7.619 & \textbf{7.506} & 9.230 & \textbf{8.915} & 9.263 & \textbf{8.789} & 10.583 & \textbf{9.841} \\
{} & {} & \small{(0.092)} & \small{(0.180)} & \small{(0.126)} & \small{(0.241)} & \small{(0.089)} & \small{(0.146)} & \small{(0.148)} & \small{(0.167)} \\

{} & \multirow{2}{*}{\NHITS} & 6.807 & \textbf{6.566} & 8.454 & \textbf{7.784} & 8.356 & \textbf{8.231} & 9.585 & \textbf{9.393} \\
{} & {} & \small{(0.050)} & \small{(0.076)} & \small{(0.212)} & \small{(0.250)} & \small{(0.020)} & \small{(0.031)} & \small{(0.080)} & \small{(0.057)} \\
\hline

\multirow{5}{*}{$L=60$ timesteps}  & \multirow{2}{*}{\TFT} & 6.087 & \textbf{5.686} & 7.344 & \textbf{6.748} & \textbf{9.012} & 9.236 & \textbf{10.090} & 10.207 \\
{} & {} & \small{(0.466)} & \small{(0.073)} & \small{(0.461)} & \small{(0.177)} & \small{(0.195)} & \small{(0.544)} & \small{(0.151)} & \small{(0.485)} \\

\multirow{3}{*}{(5 hours)}& \multirow{2}{*}{\NBEATSx} & 7.241 & \textbf{7.200} & 8.511 & \textbf{8.395} & 9.810 & \textbf{9.572} & 10.995 & \textbf{10.621} \\
{} & {} & \small{(0.052)} & \small{(0.103)} & \small{(0.064)} & \small{(0.117)} & \small{(0.122)}  & \small{(0.710)} & \small{(0.145)} & \small{(0.689)} \\

{} & \multirow{2}{*}{\NHITS} & 6.934 & \textbf{6.662} & 8.543 & \textbf{7.931} & 8.528 & \textbf{8.462} & 9.731 & \textbf{9.691} \\
{} & {} & \small{(0.050)} & \small{(0.092)} & \small{(0.151)} & \small{(0.216)} & \small{(0.046)} & \small{(0.059)} & \small{(0.089)} & \small{(0.079)}  \\
\hline

\multirow{5}{*}{$L=120$ timesteps}  & \multirow{2}{*}{\TFT} & 5.826 & \textbf{5.743} & 7.132 & \textbf{7.042} & \textbf{9.197} & 9.529 & 10.643 & \textbf{10.461} \\
{} & {} & \small{(0.192)} & \small{(0.159)} & \small{(0.243)} & \small{(0.270)} & \small{(0.455)} & \small{(0.597)} & \small{(0.792)} & \small{(0.498)} \\

\multirow{3}{*}{(10 hours)} & \multirow{2}{*}{\NBEATSx} & 6.886 & \textbf{6.806} & 8.145 &\textbf{ 7.998} & 10.280 & \textbf{9.983} & 11.359 & \textbf{10.927} \\
 & {} & \small{(0.031)} & \small{(0.060)} & \small{(0.072)} & \small{(0.076)} & \small{(0.214)} & \small{(0.264)} & \small{(0.188)} & \small{(0.271)} \\
 
{} & \multirow{2}{*}{\NHITS} & 7.304 & \textbf{7.037} & 9.094 & \textbf{8.492} & 8.873 & \textbf{8.758} & 10.137 & \textbf{9.965} \\
{} & {} & \small{(0.055)} & \small{(0.093)} & \small{(0.236)} & \small{(0.220)} & \small{(0.064)} & \small{(0.080)} & \small{(0.108)} & \small{(0.095)} \\
\hline

\multirow{5}{*}{$L=180$ timesteps}  & \multirow{2}{*}{\TFT} & 5.967 & \textbf{5.829} & 7.243 & \textbf{7.048} & 8.926 & \textbf{8.852} & 10.314 & \textbf{10.165} \\
{} & {} & \small{(0.749)} & \small{(0.122)} & \small{(0.884)} & \small{(0.257)} & \small{(0.229)} & \small{(0.195)} & \small{(0.593)} & \small{(0.400)} \\

\multirow{3}{*}{(15 hours)} & \multirow{2}{*}{\NBEATSx} & 6.803 & \textbf{6.731} & 7.900 & \textbf{7.854} & \textbf{10.356} & 10.543 & \textbf{11.493} & 11.630 \\
{} & {} & \small{(0.035)} & \small{(0.045)} & \small{(0.061)} & \small{(0.073)} & \small{(0.307)} & \small{(0.363)} & \small{(0.366)} & \small{(0.488)} \\

{} & \multirow{2}{*}{\NHITS} & 7.446 & \textbf{7.289} & 9.283 & \textbf{8.867} & \textbf{8.981} & 9.087 & \textbf{10.130} & 10.249 \\

{} & {} & \small{(0.046)} & \small{(0.040)} & \small{(0.238)} & \small{(0.122)} & \small{(0.058)} & \small{(0.131)} & \small{(0.043)} & \small{(0.128)} \\

\bottomrule
\end{tabular}}
\label{tab:contextlen_ablation_table}
\end{table*}

\begin{figure*}[t!]
    \centering
    \begin{minipage}{0.49\textwidth}
        \centering
        \includegraphics[width=\textwidth]{images/hyperglycemia_TPR_simglucose.pdf}
        (a) Simulated Dataset
    \end{minipage} \hfill
    \begin{minipage}{0.49\textwidth}
        \centering
        \includegraphics[width=\textwidth]{images/hyperglycemia_TPR_ohiot1dm.pdf}
        (b) OhioT1DM Dataset
    \end{minipage}

    \begin{minipage}{0.49\textwidth}
        \centering
        \includegraphics[width=\textwidth]{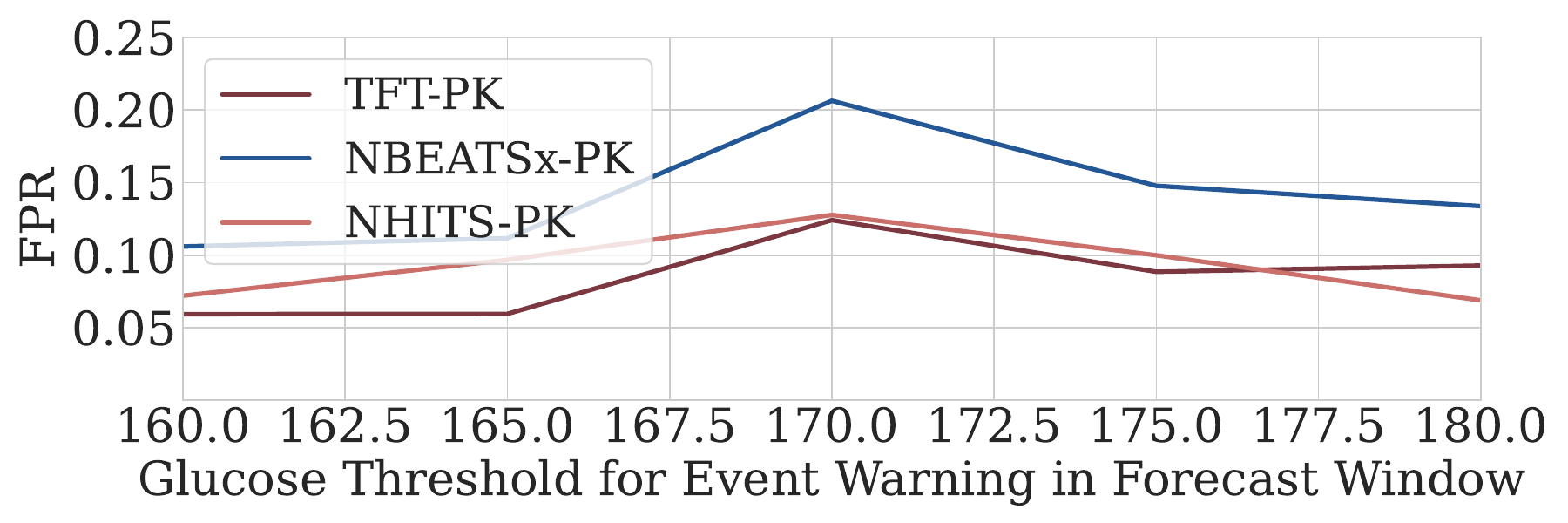}
        (c) Simulated Dataset
    \end{minipage} \hfill
    \begin{minipage}{0.49\textwidth}
        \centering
        \includegraphics[width=\textwidth]{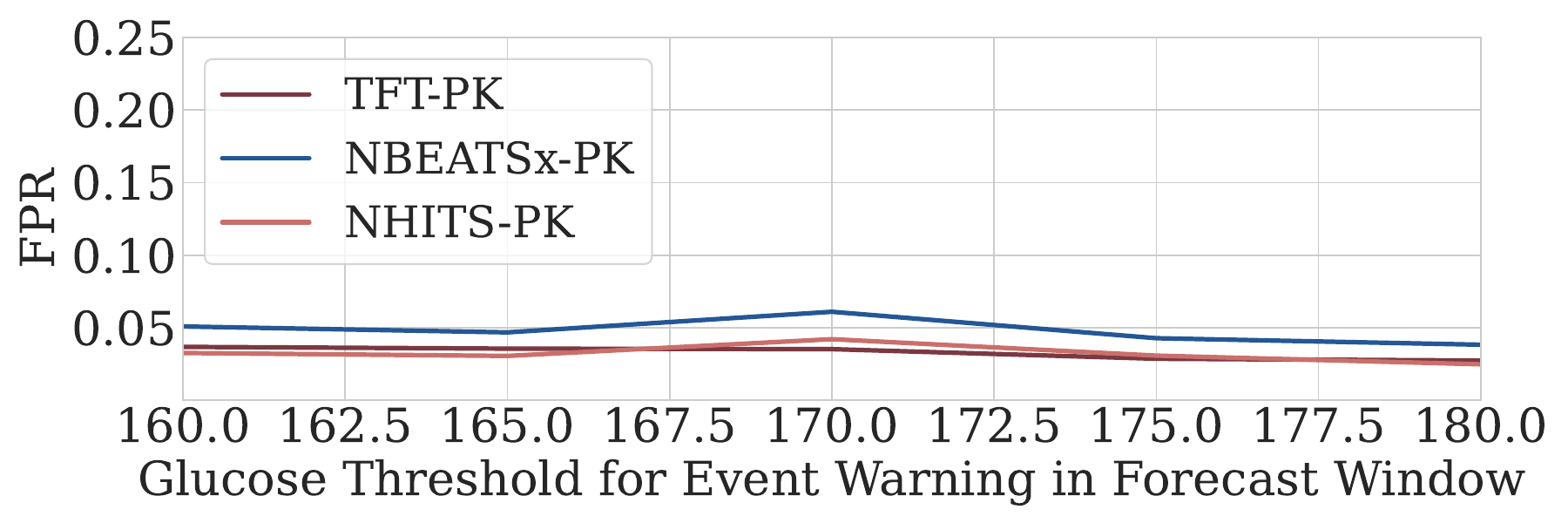}
        (d) OhioT1DM Dataset
    \end{minipage}
    \caption{\textbf{(a, b)} True positive rate (TPR) and \textbf{(c, d)} false positive rate (FPR) for the simulated and OhioT1DM datasets at various glucose thresholds for \textbf{hyperglyemia} event warnings (blood glucose $\geq 180$). TPR represents the proportion of forecast horizon windows in which the model correctly predicts an event when one actually occurs, while FPR represents the proportion of forecast horizon windows in which the model predicts an event when none occurs. The best performing \TFT-PK model for the simulated dataset predicts key glucose thresholds in forecast windows at an approximate TPR above of 0.96 and an FPR of 0.09. The best performing \NHITS-PK model for the OhioT1DM dataset predicts key glucose thresholds in forecast windows at an approximate TPR of 0.89 and an FPR of 0.02. The TPR increases as blood glucose thresholds decrease toward 160, with lower thresholds potentially serving as preliminary warning triggers to prompt patients to monitor their blood glucose and stay alert for further triggers at higher glucose levels.}
    \label{fig:hyperglycemia_tpr_fpr}
\end{figure*}

\begin{figure*}[t!]
    \centering
    \begin{minipage}{0.49\textwidth}
        \centering
        \includegraphics[width=\textwidth]{images/hypoglycemia_TPR_simglucose.pdf}
        (a) Simulated Dataset
    \end{minipage} \hfill
    \begin{minipage}{0.49\textwidth}
        \centering
        \includegraphics[width=\textwidth]{images/hypoglycemia_TPR_ohiot1dm.pdf}
        (b) OhioT1DM Dataset
    \end{minipage}

    \begin{minipage}{0.49\textwidth}
        \centering
        \includegraphics[width=\textwidth]{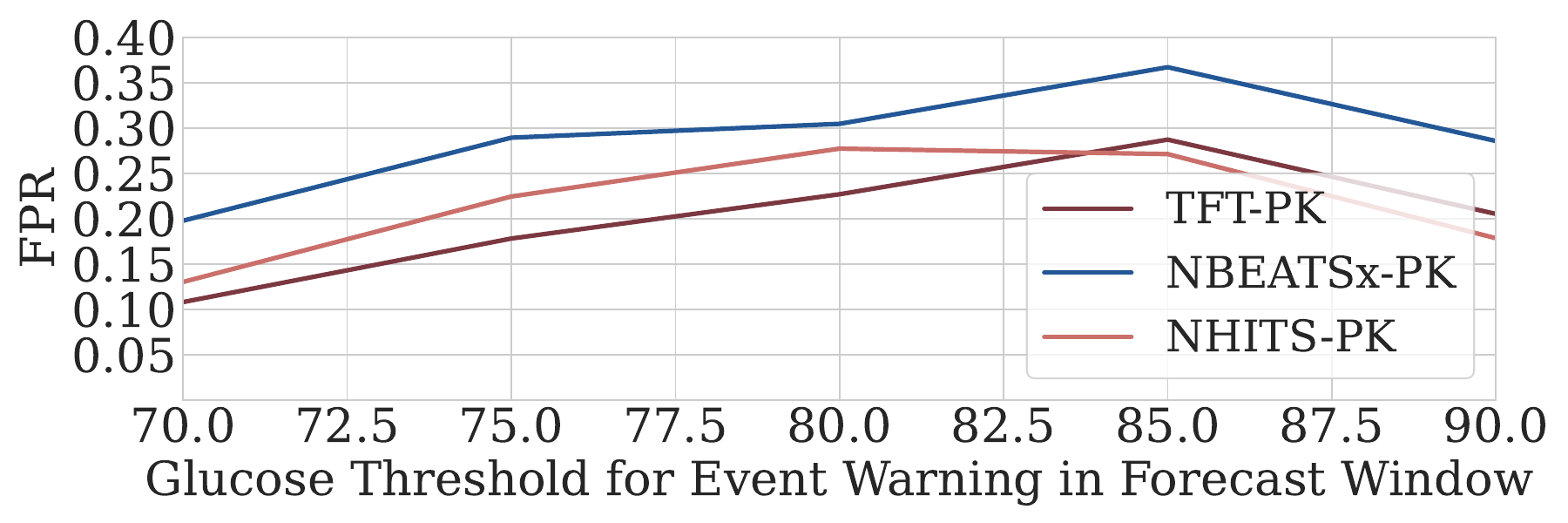}
        (c) Simulated Dataset
    \end{minipage} \hfill
    \begin{minipage}{0.49\textwidth}
        \centering
        \includegraphics[width=\textwidth]{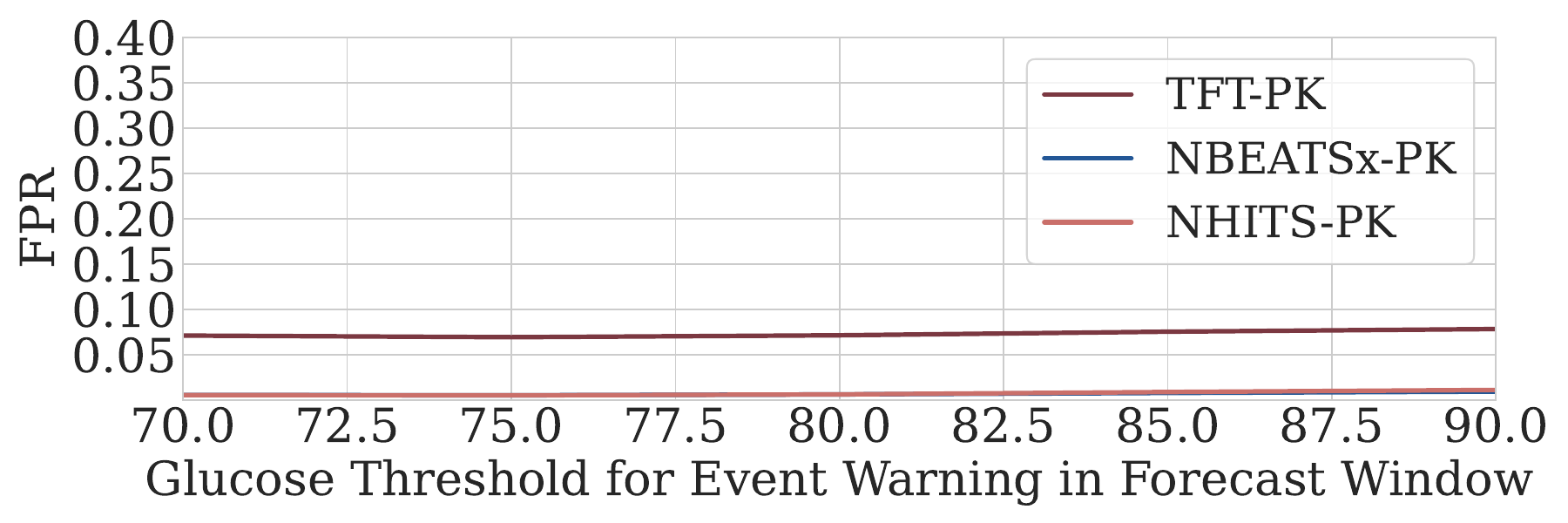}
        (d) OhioT1DM Dataset
    \end{minipage}
    \caption{\textbf{(a, b)} True positive rate (TPR) and \textbf{(c, d)} false positive rate (FPR) for the simulated and OhioT1DM datasets at various glucose thresholds for \textbf{hypoglyemia} event warnings (blood glucose $\leq 70$). TPR represents the proportion of forecast horizon windows in which the model correctly predicts an event when one actually occurs, while FPR represents the proportion of forecast horizon windows in which the model predicts an event when none occurs. The best performing \TFT-PK model for the simulated dataset predicts key glucose thresholds in forecast windows at an approximate TPR of 0.93 and an FPR of 0.11. The best performing \NHITS-PK model for the OhioT1DM dataset predicts key glucose thresholds in forecast windows at an approximate TPR of 0.69 and an FPR of 0.01. The TPR increases as blood glucose thresholds increase toward 90, with higher thresholds potentially serving as preliminary warning triggers to prompt patients to monitor their blood glucose and stay alert for further triggers at lower glucose levels.}
    \label{fig:hypoglycemia_tpr_fpr}
\end{figure*}

\subsection{Lag Time Ablation}\label{apd:contextlen_ablation}

We performed an ablation study on model performance for various lag times $L$, including 2.5 hours (30 time steps), 5 hours (60 time steps), and 10 hours (120 time steps; results presented in the main paper), as well as 15 hours (180 time steps), for the top three performing deep learning models: \TFT, \NBEATSx, and \NHITS. The study was conducted at a fixed forecast horizon $H$ of 30 minutes (6 time steps). For each lag time, the \NBEATSx-PK and \NHITS-PK models had a lower MAE than their sparse exogenous model counterparts, except for $L\!=\!180$ in the OhioT1DM dataset, as shown in Table~\ref{tab:contextlen_ablation_table}. The \TFT-PK model had a lower MAE than its sparse exogenous model counterpart for $L \in \{60, 120, 180\}$ in the simulated dataset and for $L\!=\!180$ in the OhioT1DM dataset.

For each lag time, we evaluate whether the PK models have a significantly lower forecasting error, in terms of MAE, than their sparse exogenous model counterparts using paired t-tests, as outlined in Section~\ref{apd:evaluation}. For the simulated dataset, the \NBEATSx-PK and \NHITS-PK models showed a statistically significant performance improvement over their sparse exogenous model counterparts ($\text{p-value}<0.05$) for all lag times, while the \TFT-PK model showed a statistically significant performance improvement for all lag times except $L=2.5$. For the OhioT1DM dataset, the \NBEATSx-PK and \NHITS-PK models demonstrated a statistically significant performance improvement over their sparse exogenous model counterparts ($\text{p-value}<0.05$) for lag times of 2.5, 5, and 10 hours, while the \TFT-PK model showed a statistically significant performance improvement for $L\!=\!180$. Notably, for $L\!=\!180$, the \TFT-PK model was also the best-performing \TFT\ model variant across all lag times for the OhioT1DM dataset.

\subsection{Critical Event Prediction Rate Analysis}\label{apd:tpr_fpr_analysis}

We evaluate the ability of models to predict critical events using two metrics: the true positive rate (TPR), or the proportion of forecast horizon windows in which the model correctly predicts an event when one actually occurs, and the false positive rate (FPR), or the proportion of forecast horizon windows in which the model predicts an event when none occurs. From the TPR, one can also obtain the false negative rate ($\text{FNR} = 1 - \text{TPR}$), or the proportion of forecast horizon windows in which the model does not predict an event when one actually occurs. We measure each metric on multiple blood glucose thresholds up to critical event thresholds (blood glucose $\leq 70 | \geq 180$).

The TPR and FPR for the simulated and OhioT1DM datasets at various glucose thresholds for hyperglycemia event warnings are shown in Fig.~\ref{fig:hyperglycemia_tpr_fpr}. The best performing \TFT-PK model for the simulated dataset predicts hyperglycemic events (blood glucose $\geq 180$) in forecast windows at an approximate TPR of 0.96 and an FPR of 0.09. The best performing \NHITS-PK model for the OhioT1DM dataset predicts hyperglycemic events in forecast windows at an approximate TPR of 0.89 and an FPR of 0.02. The TPR increases as blood glucose thresholds decrease toward 160 mg/dL, with lower thresholds potentially serving as preliminary warnings, prompting patients to monitor their blood glucose and stay alert for further warnings at higher glucose levels.

The TPR and FPR for the simulated and OhioT1DM datasets at various glucose thresholds for hypoglycemia event warnings are shown in Fig.~\ref{fig:hypoglycemia_tpr_fpr}. The best performing \TFT-PK model for the simulated dataset predicts hypoglycemic events (blood glucose $\leq 70$) in forecast windows at an approximate TPR of 0.93 and an FPR of 0.11. The best performing \NHITS-PK model for the OhioT1DM dataset predicts hypoglycemic events in forecast windows at an approximate TPR of 0.69 and an FPR of 0.01. The TPR increases as blood glucose thresholds decrease toward 90 mg/dL, with higher thresholds potentially serving as preliminary warnings to prompt patients to monitor their blood glucose and stay alert for further warnings at lower glucose levels.

These results highlight the effectiveness of the models in predicting critical glucose events, demonstrating high TPRs while maintaining relatively low FPRs. The ability to provide early warnings at varying glucose thresholds can be valuable for proactive patient monitoring and timely intervention. However, there is still room for improvement, as achieving higher TPRs with even lower FPRs could further enhance reliability and reduce false alarms.

\subsection{Computational Efficiency Analysis}\label{apd:computational_efficiancy_analysis}

\begin{table*}[t!]
\centering
\begin{tabular}{l|l|l|c}
\hline
\textbf{Model} & \textbf{FLOPS} & \textbf{\# Trainable Parameters} & \textbf{Inference Time (ms)} \\
\hline
\TFT\ w/o Exog. & 3.66e8 & 2.53 million & 15.9 \\
\TFT\ w/ Exog. & 6.11e8 ($\uparrow 66.94\%$) & 2.78 million ($\uparrow 9.88\%$) & 17.7 \\
\TFT-PK w/ Exog. & 6.07e8 ($\uparrow 65.85\%$) & 2.78 million ($\uparrow 9.88\%$) & 21.3 \\
\hline
\NBEATSx\ w/o Exog. & 8.01e7 & 10.02 million & 5.2 \\
\NBEATSx\ w/ Exog. & 8.89e7 ($\uparrow 10.99\%$) & 11.13 million ($\uparrow 11.08\%$) & 5.8 \\
\NBEATSx-PK w/ Exog. & 8.30e7 ($\uparrow 3.62\%$) & 10.39 million ($\uparrow 3.69\%$)& 5.7 \\
\hline
\NHITS\ w/o Exog. & 8.19e7 & 10.25 million & 5.5 \\
\NHITS\ w/ Exog. & 9.08e7 ($\uparrow 10.87\%$) & 11.36 million ($\uparrow 10.83\%$) & 5.8 \\
\NHITS-PK w/ Exog. & 8.49e7 ($\uparrow 3.66\%$) & 10.62 million ($\uparrow 3.61\%$) & 8.4 \\
\hline
\end{tabular}
\caption{Computational efficiency characteristics including FLOPS (floating-point operations per second), number of trainable parameters, and inference time measured for the OhioT1DM dataset models. The percentage increase in FLOPs and the number of trainable parameters over the baseline model without (w/o) exogenous (exog.) variables is shown in parentheses (e.g., $\uparrow \text{x.xx}\%$). As expected, models with (w/) exogenous variables exhibit larger FLOPs, more trainable parameters, and longer inference times due to the processing of higher-dimensional input. However, the PK models have the same or fewer FLOPs and trainable parameters compared to their sparse exogenous model counterparts, as the exogenous features are leveraged more effectively in the PK model architectures.}
\label{tab:efficiency_metrics}
\end{table*}

To assess the computational efficiency characteristics of the deep learning models, we calculate three key metrics: FLOPs (floating-point operations per second), the number of trainable parameters, and the inference time for the OhioT1DM dataset models. These metrics provide insights into the computational complexity, memory usage, and real-time efficiency of the model.
We calculate FLOPs using the FlopCounterMode class from Pytorch, which tracks operations during model execution. The resulting FLOPs represent the computational load of the model for each forward pass. The number of trainable parameters in a model reflects its size. We calculate this by iterating through all the model's parameters and summing the number of elements for those that require gradients. Inference time refers to the time it takes for the model to generate predictions from input data. To measure this, we execute the model multiple times (100 runs) and record the time taken for each forward pass. The inference time is then calculated as the average time taken across these runs, providing an estimate of the model's efficiency during real-time deployment. This metric is essential for applications like clinical decision support or real-time monitoring devices, where timely predictions are crucial. The metrics for the three best performing deep learning models, \TFT-PK, \NBEATSx-PK, and \NHITS-PK are presented in Table~\ref{tab:efficiency_metrics}. In addition to providing metrics for the PK models, the table also includes metrics for models with (w/) and without (w/o) exogenous (exog.) features for reference.
As expected, models with exogenous variables exhibit larger FLOPs, more trainable parameters, and longer inference times due to the processing of higher-dimensional input. However, the PK models have the same or fewer FLOPs and trainable parameters compared to their sparse exogenous model counterparts, as the exogenous features are leveraged more effectively in the PK model architectures. For \NBEATSx\ and \NHITS\ models, this efficiency stems from our proposed approach, where exogenous variables are processed in only one of the deep stacks of fully connected layers rather than across all stacks. This design improves processing efficiency and reduces model size. Without the PK encoder, the approach of isolating sparse exogenous variables to a single stack generally results in worse performance, as shown in Table~\ref{table:single_stack_exp}.

The \TFT-PK, \NBEATSx-PK, and \NHITS-PK models with sparse exogenous features show increases in FLOP of 65.30\%, 3.62\%, and 3.66\%, respectively, compared to the baseline models without exogenous variables. In contrast, the models with sparse exogenous features exhibit higher FLOP increases of 66.94\%, 10.99\%, and 10.87\%, respectively, compared to the baseline models without exogenous variables. This result highlights the utility of the PK models, as they not only provide lower forecasting errors but also offer greater computational efficiency. Additionally, the models have fast inference speeds, on the order of magnitudes smaller than 1 second.

\begin{table}[t!]
\caption{Processing covariates in 1 versus 3 of the deep stacks of fully connected layers for \NBEATSx\ and \NHITS. Without the PK encoder, the approach of isolating sparse exogenous variables to a single stack generally results in worse performance.}
\centering
\resizebox{0.495\textwidth}{!}{\begin{tabular}{l|cc|cc}
\toprule
{} & \multicolumn{2}{c|}{\textbf{Simulated Dataset}} & \multicolumn{2}{c}{\textbf{OhioT1DM Dataset}} \\
{} & \textbf{\NBEATSx} & \textbf{\NHITS} & \textbf{\NBEATSx} & \textbf{\NHITS} \\
\hline
\multirow{2}{*}{1 Stack} & 6.908 & 7.429 & \textbf{10.020} & 8.897 \\
{} & \small{(0.101)} & \small{(0.042)} & \small{(0.569)} & \small{(0.246)} \\
\hline
\multirow{2}{*}{3 Stacks} & \textbf{6.886} & \textbf{7.304} & 10.280 & \textbf{8.873} \\
{} & \small{(0.031)} & \small{(0.055)} & \small{(0.214)} & \small{(0.064)} \\
\bottomrule
\end{tabular}}
\label{table:single_stack_exp}
\end{table}

\subsection{Feasibility of Model Integration into Clinical Systems}\label{apd:clinical_system_feasibility}
Integrating deep learning forecasting models into clinical decision-support systems or real-time monitoring devices can provide timely, data-driven alerts for critical events, aiding decision-making for proactive interventions. To ensure smooth integration into a clinical decision-support system or real-time monitoring device, deep learning models must be both data-efficient and capable of delivering reasonable inference times to facilitate updated forecasts at appropriate time intervals. The OhioT1DM patients use Medtronic 530G or 630G insulin pumps and Medtronic Enlite Continuous Glucose Monitor (CGM) sensors. The MiniMed 630G Insulin Pump is capable of storing 90 days of pump history and glucose sensor data. Given that such devices are equipped to store data on the scale of many days, the feasibility of integrating the proposed hybrid global-local PK models into such systems is high, as our PK models only require a maximum of 10 hours of historical data to generate forecasts. The PK models are also memory-efficient. For example, the \NHITS-PK model requires less than 20MB of memory during inference on a CPU in a cold-start scenario. A cold start refers to the initial model inference, where no preloaded state or cached data is used, providing an upper bound on memory usage. Furthermore, the \NHITS-PK model demonstrates an inference time of approximately 8 milliseconds when executed on a CPU, ensuring rapid predictions that support real-time operation. This combination of low storage requirements and fast inference times makes the integration of such models into clinical systems feasible to help monitor and provide early warnings for critical events. While a graphics processing unit (GPU) is recommended for efficient model training, a central processing unit (CPU) is sufficient for inference with the trained model, offering space, cost, and energy efficiency benefits over dedicated graphics processing devices.

In addition to CGMs, deep learning models could be integrated into other mobile devices, such as a phone app. The model could provide continuous forecast windows that update every 5 minutes (the frequency of the data) and trigger a beep or vibration if the patient is at risk of a critical event occurring within the forecast horizon. The device could alert the patient with customizable blood glucose thresholds, accounting for warnings of varying severity. For example, preliminary warnings could be set at less severe blood glucose levels to prompt patients to monitor their blood glucose and remain alert for further triggers at more severe thresholds. Furthermore, integrating the model into a phone app could facilitate sharing alert information, notifying patients and allowing the inclusion of other monitoring users, such as parents, school nurses, or caregivers.

\begin{table*}[ht!]
\caption{Mean absolute error (MAE) computed for model predictions across all values in the forecast horizon and only across critical values in the forecast horizon (blood glucose $\leq 70 | \geq 180$) for simulated dataset and OhioT1DM dataset. Average results are computed across 8 trials.  Models are separated with horizontal lines into statistical, linear, RNN-based, CNN-based, Transformer-based, and MLP-based, and PK model groups, respectively. Result for models with (w/) and without (w/o) exogenous (exog.) variables are shown where applicable. Best results are highlighted in \textbf{bold}, second best results are \underline{underlined}. PK models with improved forecasts over their respective exogenous baseline model are highlighted in \textcolor{blue}{blue}.}
\centering
\resizebox{1.0\textwidth}{!}{\begin{tabular}{l|cc cc|cc cc}
\toprule
{} & \multicolumn{4}{c|}{\textbf{Simulated Dataset}} & \multicolumn{4}{c}{\textbf{OhioT1DM Dataset}} \\

\multirow{2}{*}{\textbf{Model}} & \multicolumn{2}{c}{\textbf{\small{All Values}}} & \multicolumn{2}{c|}{\textbf{\small{Critical Values}}} & \multicolumn{2}{c}{\textbf{\small{All Values}}} & \multicolumn{2}{c}{\textbf{\small{Critical Values}}} \\

{} & \thead{\textbf{w/o Exog.}} & \thead{\textbf{w/ Exog.}} & \thead{\textbf{w/o Exog.}} & \thead{\textbf{w/ Exog.}} & \thead{\textbf{w/o Exog.}} & \thead{\textbf{w/ Exog.}} & \thead{\textbf{w/o Exog.}} & \thead{\textbf{w/ Exog.}} \\
\hline

\multirow{2}{*}{\ETS} & 9.450 & \multirow{2}{*}{--} & 13.457 & \multirow{2}{*}{--} & 9.022 & \multirow{2}{*}{--} & \underline{10.028} & \multirow{2}{*}{--}  \\
{} & \small{(0.00)} & {} & \small{(0.00)} & {} & \small{(0.00)} & {} & \small{(0.00)} & {} \\
\hline

\multirow{2}{*}{\DLinear} & 7.802 & 7.989 & 10.575 & 10.790 & 11.514 & 13.011 & 12.512 & 14.580 \\
{} & \small{(0.054)} & \small{(0.141)} & \small{(0.126)} & \small{(0.169)} & \small{(1.595)} & \small{(3.590)} & \small{(1.228)} & \small{(4.080)} \\
\hline

\multirow{2}{*}{\RNN} & 12.728 & 12.930 & 20.308 & 20.945 & 11.535 & 10.907 & 12.641 & 12.989  \\
{} & \small{(8.829)} & \small{(8.819)} & \small{(19.533)} & \small{(19.925)} & \small{(3.056)} & \small{(0.809)} & \small{(2.342)} & \small{(1.822)} \\ 

\multirow{2}{*}{\LSTM} & 9.250 & 9.152 & 12.034 & 12.000 & 10.217 & 10.615 & 12.102 & 12.983  \\
{} & \small{(2.041)} & \small{(2.097)} & \small{(1.830)} & \small{(3.983)} & \small{(0.930)} & \small{(0.807)} & \small{(1.218)} & \small{(1.464)} \\
\hline

\multirow{2}{*}{\TCN} & 14.340 & 11.380 & 23.370 & 14.817 & 10.769 & 11.516 & 12.812 & 13.203 \\
{} & \small{(8.271)} & \small{(2.049)} & \small{(20.857)} & \small{(3.258)} & \small{(1.093)} & \small{(1.846)} & \small{(0.901)} & \small{(1.882)} \\
\hline

\multirow{2}{*}{\Informer} & 8.599 & \multirow{2}{*}{--} & 12.074 & \multirow{2}{*}{--} & 13.714 & \multirow{2}{*}{--} & 14.867 & \multirow{2}{*}{--}  \\
{} & \small{(1.060)} & {} & \small{(2.083)} & {} & \small{(3.753)} & {} & \small{(3.620)} & {} \\

\multirow{2}{*}{\PatchTST} & 7.284 & \multirow{2}{*}{--} & 9.842 & \multirow{2}{*}{--} & 9.922 & \multirow{2}{*}{--} & 10.999 & \multirow{2}{*}{--} \\
{} & \small{(0.554)} & {} & \small{(0.939)} & {} & \small{(0.949)} & {}& \small{(1.054)} & {} \\

\multirow{2}{*}{\TFT} & 6.587 & \underline{5.826} & 8.518 & \underline{7.132} & 9.402 & 9.197 & 10.577 & 10.643  \\
{} & \small{(0.115)} & \small{(0.192)} & \small{(0.178)} & \small{(0.243)} & \small{(0.811)} & \small{(0.455)} & \small{(1.038)} & \small{(0.792)} \\

\hline
\multirow{2}{*}{\MLP} & 10.424 & 8.407 & 13.932 & 10.297 & 10.300 & 13.254 & 11.550 & 14.997 \\
{} & \small{(1.478)} & \small{(0.387)} & \small{(0.861)} & \small{(0.329)} & \small{(0.329)} & \small{(0.245)} & \small{(0.359)} & \small{(0.353)} \\

\multirow{2}{*}{\NBEATSx} & 9.187 & 6.886 & 12.757 & 8.145 & 9.868 & 10.280 & 11.037 & 11.359 \\
{} & \small{(0.236)} & \small{(0.031)} & \small{(0.547)} & \small{(0.072)} & \small{(0.528)} & \small{(0.214)} & \small{(0.548)} & \small{(0.188)} \\

\multirow{2}{*}{\NHITS} & 8.268 & 7.304 & 11.114 & 9.094 & 9.060 & \underline{8.873} & 10.337 & 10.137 \\
{} & \small{(0.039)} & \small{(0.055)} & \small{(0.237)} & \small{(0.236)} & \small{(0.142)} & \small{(0.064)} & \small{(0.137)} & \small{(0.108)} \\
\hline

\multirow{2}{*}{\TFT-PK} & \multirow{2}{*}{--} & \textcolor{blue}{\textbf{5.743}} & \multirow{2}{*}{--} & \textcolor{blue}{\textbf{7.042}} & \multirow{2}{*}{--} & 9.529 & \multirow{2}{*}{--} & \textcolor{blue}{10.461} \\
{} & {} & \small{(0.159)} & {} & \small{(0.270)} & {} & \small{(0.597)} & {} & \small{(0.498)} \\

\multirow{2}{*}{\NBEATSx-PK} & \multirow{2}{*}{--} & \textcolor{blue}{6.806} & \multirow{2}{*}{--} & \textcolor{blue}{7.998} & \multirow{2}{*}{--} & \textcolor{blue}{9.983} & \multirow{2}{*}{--} & \textcolor{blue}{10.927} \\
{} & {} & \small{(0.060)} & {} & \small{(0.076)} & {} & \small{(0.264)} & {} & \small{(0.271)} \\

\multirow{2}{*}{\NHITS-PK} & \multirow{2}{*}{--} & \textcolor{blue}{7.037} & \multirow{2}{*}{--} & \textcolor{blue}{8.492} & \multirow{2}{*}{--} & \textcolor{blue}{\textbf{8.758}} & \multirow{2}{*}{--} & \textcolor{blue}{\textbf{9.965}}  \\
{} & {} & \small{(0.093)} & {} & \small{(0.220)} & {} & \small{(0.080)} & {} & \small{(0.095)} \\
\bottomrule
\end{tabular}}
\label{mae_table_appendix}
\end{table*}

\begin{table*}[t!]
\caption{Root mean square error (RMSE) computed for model predictions across all values in the forecast horizon and only across critical values in the forecast horizon (blood glucose $\leq 70 | \geq 180$) for simulated dataset and OhioT1DM dataset. Average results are computed across 8 trials. Models are separated with horizontal lines into statistical, linear, RNN-based, CNN-based, Transformer-based, and MLP-based groups, and PK model groups, respectively. Result for models with (w/) and without (w/o) exogenous (exog.) variables are shown where applicable. Best results are highlighted in \textbf{bold}, second best results are \underline{underlined}. PK models with improved forecasts over their respective exogenous baseline model are highlighted in \textcolor{blue}{blue}.}
\centering
\resizebox{1.0\textwidth}{!}{\begin{tabular}{l|cc cc|cc cc}
\toprule
{} & \multicolumn{4}{c|}{\textbf{Simulated Dataset}} & \multicolumn{4}{c}{\textbf{OhioT1DM Dataset}}
\\
\multirow{2}{*}{\textbf{Model}} & \multicolumn{2}{c}{\textbf{\small{All Values}}} & \multicolumn{2}{c|}{\textbf{\small{Critical Values}}} & \multicolumn{2}{c}{\textbf{\small{All Values}}} & \multicolumn{2}{c}{\textbf{\small{Critical Values}}}  
\\
{} & \thead{\textbf{w/o Exog.}} & \thead{\textbf{w/ Exog.}} & \thead{\textbf{w/o Exog.}} & \thead{\textbf{w/ Exog.}} & \thead{\textbf{w/o Exog.}} & \thead{\textbf{w/ Exog.}} & \thead{\textbf{w/o Exog.}} & \thead{\textbf{w/ Exog.}} \\
\hline

\multirow{2}{*}{\ETS} & 14.751 & \multirow{2}{*}{--} & 21.454 & \multirow{2}{*}{--} & 14.569 & \multirow{2}{*}{--} & \textbf{16.225} & \multirow{2}{*}{--}  \\
{} & \small{(0.00)} & {} & \small{(0.00)} & {} & \small{(0.00)} & {} & \small{(0.00)} & {} \\
\hline

\multirow{2}{*}{\DLinear} & 12.545 & 12.673 & 17.606 & 17.707 & 17.138 & 18.609 & 18.679 & 20.743 \\
{} & \small{(0.085)} & \small{(0.119)} & \small{(0.340)} & \small{(0.216)} & \small{(1.462)} & \small{(3.511)} & \small{(0.970)} & \small{(3.869)} \\
\hline

\multirow{2}{*}{\RNN} & 17.679 & 18.017 & 26.312 & 26.849 & 16.433 & 15.762 & 18.274 & 18.375 \\
{} & \small{(10.978)} & \small{(10.935)} & \small{(19.591)} & \small{(19.935)} & \small{(2.734)} & \small{(0.657)} & \small{(2.199)} & \small{(1.654)} \\ 

\multirow{2}{*}{\LSTM} & 12.993 & 12.526 & 17.027 & 16.222 & 15.337 & 15.571 & 17.759 & 18.399  \\
{} & \small{(1.896)} & \small{(2.389)} & \small{(1.599)} & \small{(4.528)} & \small{(0.750)} & \small{(0.684)} & \small{(0.949)} & \small{(1.295)} \\
\hline

\multirow{2}{*}{\TCN} & 20.040 & 15.181 & 30.812 & 19.573 & 16.073 & 16.483 & 18.642 & 18.677 \\
{} & \small{(10.281)} & \small{(2.571)} & \small{(20.017)} & \small{(4.054)} & \small{(0.855)} & \small{(1.544)} & \small{(0.601)} & \small{(1.451)} \\
\hline

\multirow{2}{*}{\Informer} & 12.348 & \multirow{2}{*}{--} & 17.400 & \multirow{2}{*}{--} & 18.536 & \multirow{2}{*}{--} & 19.904 & \multirow{2}{*}{--}  \\
{} & \small{(0.918)} & {} & \small{(1.865)} & {} & \small{(3.459)} & {} & \small{(3.220)} & {} \\

\multirow{2}{*}{\PatchTST} & 11.467 & \multirow{2}{*}{--} & 15.755 & \multirow{2}{*}{--} & 15.375 & \multirow{2}{*}{--} & 17.065 & \multirow{2}{*}{--} \\
{} & \small{(0.603)} & {} & \small{(1.000)} & {} & \small{(0.838)} & {}& \small{(0.928)} & {} \\

\multirow{2}{*}{\TFT} & 10.691 & \underline{8.898} & 14.188 & \underline{10.979} & 14.832 & 14.571 & 16.523 & 16.577 \\
{} & \small{(0.139)} & \small{(0.314)} & \small{(0.329)} & \small{(0.446)} & \small{(0.791)} & \small{(0.345)} & \small{(0.959)} & \small{(0.684)} \\
\hline

\multirow{2}{*}{\MLP} & 15.196 & 11.267 & 21.246 & 13.832 & 15.908 & 18.986 & 17.762 & 21.340 \\
{} & \small{(1.322)} & \small{(0.405)} & \small{(0.692)} & \small{(0.336)} & \small{(0.276)} & \small{(0.312)} & \small{(0.421)} & \small{(0.440)} \\

\multirow{2}{*}{\NBEATSx} & 14.105 & 9.570 & 20.526 & 11.397 & 15.934 & 16.224 & 17.540 & 17.742 \\
{} & \small{(0.348)} & \small{(0.053)} & \small{(0.756)} & \small{(0.104)} & \small{(0.548)} & \small{(0.675)} & \small{(0.599)} & \small{(0.303)} \\

\multirow{2}{*}{\NHITS} & 12.918 & 10.546 & 18.436 & 13.371 & 14.844 & \underline{14.505} & 16.789 & 16.453 \\
{} & \small{(0.064)} & \small{(0.106)} & \small{(0.369)} & \small{(0.387)} & \small{(0.192)} & \small{(0.096)} & \small{(0.189)} & \small{(0.132)} \\
\hline

\multirow{2}{*}{\TFT-PK} & \multirow{2}{*}{--} & \textcolor{blue}{\textbf{8.620}} & \multirow{2}{*}{--} & \textcolor{blue}{\textbf{10.517}} & \multirow{2}{*}{--} & 14.725 & \multirow{2}{*}{--} & \textcolor{blue}{16.265} \\
{} & {} & \small{(0.188)} & {} & \small{(0.367)} & {} & \small{(0.414)} & {} & \small{(0.437)} \\
\multirow{2}{*}{\NBEATSx-PK} & \multirow{2}{*}{--} & \textcolor{blue}{9.480} & \multirow{2}{*}{--} & \textcolor{blue}{11.244} & \multirow{2}{*}{--} & 16.945 & \multirow{2}{*}{--} & \textcolor{blue}{17.659} \\
{} & {} & \small{(0.084)} & {} & \small{(0.117)} & {} & \small{(1.467)} & {} & \small{(0.658)} \\
\multirow{2}{*}{\NHITS-PK} & \multirow{2}{*}{--} & \textcolor{blue}{10.039} & \multirow{2}{*}{--} & \textcolor{blue}{12.339} & \multirow{2}{*}{--} & \textcolor{blue}{\textbf{14.385}} & \multirow{2}{*}{--} & \textcolor{blue}{\underline{16.251}}  \\
{} & {} & \small{(0.188)} & {} & \small{(0.435)} & {} & \small{(0.092)} & {} & \small{(0.089)} \\
\bottomrule
\end{tabular}}
\label{rmse_table_appendix}
\end{table*}

\subsection{Model MAE and RMSE Results}\label{apd:second_mae_rmse}
This appendix presents the complete results for all models for both simulated data and OhioT1DM. Standard deviation across 8 trials are provided in parenthesis. Tables~\ref{mae_table_appendix} and~\ref{rmse_table_appendix} provide complete MAE and RMSE results, respectively.

\end{document}